\documentclass[final,5p,times]{elsarticle}

\usepackage{mathrsfs}
\usepackage{amsmath}
\usepackage{amssymb,amsmath}
\usepackage{algorithmic}
\usepackage{algorithm}
\usepackage{url,subfigure,epsfig,graphicx}
\usepackage{float}
\usepackage{lipsum}% <- For dummy text
\usepackage{multicol}
\usepackage{booktabs}
\usepackage{multirow}
\usepackage{rotating}
\usepackage{gensymb}
\usepackage{scalerel}
\usepackage[colorlinks,linkcolor=black,anchorcolor=black,citecolor=black]{hyperref}
\usepackage{soul}
\usepackage{color}
\usepackage{changes}
\usepackage{makecell}
\definechangesauthor[name={Zhiling Jin}, color=red]{zl}

\usepackage[switch]{lineno}

\begin{document}
% \linenumbers

\begin{frontmatter}


%% use the cortext command for theassociat  ed footnote;

\title{Dynamic Graph Neural Network with Adaptive Edge Attributes for Air Quality Prediction}


\author[1]{Jing Xu}
%\ead{xujingsss@mail.bnu.edu.cn}
\author[1,5]{Shuo Wang}
%\ead{wangshuo.sss@mail.bnu.edu.cn}
\author[2]{Na Ying}
%\ead{yingna@craes.org.cn}
\author[3]{Xiao Xiao}
%\ead{xiaoxiao@xidian.edu.cn}
\author[1,5]{Jiang Zhang}
%\ead{zhangjiang@bnu.edu.cn}
\author[4]{Yun Cheng}
%\ead{chengyu@ethz.ch}
\author[3]{Zhiling Jin}
\author[1,6]{Gangfeng Zhang\corref{cor1}}
\ead{zhanggf15@foxmail.com}

\affiliation[1]{organization={School of Systems Science, Beijing Normal University},%Department and Organization
            city={Beijing},
            postcode={100875}, 
            country={China}
            }
\affiliation[2]{organization={Chinese Research Academy of Environmental Sciences},%Department and Organization 
            city={Beijing},
            postcode={100085}, 
            country={China}
            }       
\affiliation[3]{organization={School of Telecommunications Engineering, Xidian University},%Department and Organization 
            city={Xi'an},
            postcode={710071}, 
            state={Shaanxi},
            country={China}
            }  
\affiliation[4]{organization={Information Technology and Electrical Engineering, ETH Zurich},%Department and Organization 
            city={Zurich},
            postcode={8092}, 
            country={Switzerland}
            }  
\affiliation[5]{organization={Swarma Research},%Department and Organization 
            city={Beijing},
            country={China}
            }
\affiliation[6]{organization={State Key Laboratory of Earth Surface Processes and Resource Ecology, Beijing Normal University},%Department and Organization 
            city={Beijing},
            postcode={100875}, 
            country={China}
            }  
            
\cortext[cor1]{corresponding author.}

\begin{abstract}
Air quality prediction is a typical Spatio-temporal modeling problem, which always uses different components to handle spatial and temporal dependencies in complex systems separately. Previous models based on time series analysis and Recurrent Neural Network (RNN) methods have only modeled time series while ignoring spatial information. Previous GCNs-based methods usually require providing spatial correlation graph structure of observation sites in advance. The correlations among these sites and their strengths are usually calculated using prior information. However, due to the limitations of human cognition, limited prior information cannot reflect the real station-related structure or bring more effective information for accurate prediction. To this end, we propose a novel Dynamic Graph Neural Network with Adaptive Edge Attributes (DGN-AEA) on the message passing network, which generates the adaptive bidirected dynamic graph by learning the edge attributes as model parameters. Unlike prior information to establish edges, our method can obtain adaptive edge information through end-to-end training without any prior information. Thus reduced the complexity of the problem. Besides, the hidden structural information between the stations can be obtained as model by-products, which can help make some subsequent decision-making analyses. Experimental results show that our model received state-of-the-art performance than other baselines.
\end{abstract}

% \includegraphics[width=18cm]{figures/framework-new.pdf}

% \begin{highlights}
%   \item We develop an adaptive dynamic graph learning unit to learn the graph automatically.
%   \item The wind field data is integrated as a type of directed dynamic connection.
%   \item Non-Euclidean regional modeling without prior information can be solved.
%   \item Our model can be seen as dynamic multi-graph structure compared with others. 
%   \item The proposed DGN-AEA model performs at the state-of-the-art level.
% \end{highlights}

\begin{keyword}


Air quality prediction \sep Adaptive graph learning \sep Dynamic Graph \sep Message Passing Neural Networks
\end{keyword}

\end{frontmatter}


\section{Introduction}
\label{sec:introduction}
Air quality has a significant impact on our daily lives. People who breathe clean air sleep better and are less likely to die prematurely from diseases such as cardiovascular and respiratory disorders, as well as lung cancer \cite{strom2016effects,vandyck2018air}. One of the key factors that decreases air quality is $\mathrm{PM}_{2.5}$ (atmospheric particulate matter (PM) having a diameter of 2.5 $\mu \mathrm{m}$ or less), which can easily be inhaled and cause damage to the human body. Thus, monitoring and forecasting the $\mathrm{PM}_{2.5}$ concentrations are critical for improving air quality. With the rapid development of industry, a significant amount of energy is consumed, resulting in massive $\mathrm{PM}_{2.5}$ emissions \cite{zhang2016air}. According to \cite{li2018evaluation}, in 2016, 38 days were heavily polluted due to $\mathrm{PM}_{2.5}$ emissions in Beijing. High $\mathrm{PM}_{2.5}$ concentrations may cause serious adverse health impacts and diseases \cite{lu2019analysis}, such as cardiac and pulmonary disease \cite{leclercq2018air}, detrimental effects on birth outcomes \cite{sun2020prenatal}, and infant mortality \cite{li2021effect}. Fortunately, to monitor and record air quality data, a large number of low-cost air quality sensors have been deployed, which makes it possible for researchers to perform accurate air quality prediction tasks. Accurate air quality predictions are useful, for example, individual activity arrangements and government pollution restrictions can benefit from that \cite{yi2018deep}. When the predicted $\mathrm{PM}_{2.5}$ concentrations are too high, people can avoid going out and politicians can modify policies accordingly. 

Conventional air quality prediction approaches can be generally divided into two categories: the knowledge-driven approach and the data-driven approach. In the past two decades, the knowledge-driven approach has been widely adopted for air quality prediction. The representations of this approach include the community multiscale air quality (CMAQ) \cite{murphy2020detailed}, comprehensive air quality model with extensions (CAMx) \cite{ibrahim2019air}, weather research, and forecasting/chemistry (WRF-Chem) \cite{hong2020improved} modeling systems. This knowledge-driven approach does not necessitate a large number of historical observations, and its prediction performance is determined by how well the model fits the actual conditions. However, in the real world, air quality data change dramatically, which makes it not easy to make this model fit the conditions. In addition, the air quality is affected by various features, such as the weather, altitude, and the wind field. Under these circumstances, the knowledge-driven approaches often do not yield good predictions.

Recently, data-driven approaches have shown great performance for air quality prediction. As a conventional data-driven approach, statistical methods have been widely adopted for their simple structure. Yi et al. \cite{yi2018deep} applied the autoregressive integrated moving average (ARIMA) model to capture the trend of air quality time series in New Delhi. Naveen et al. \cite{naveen2017time} then adopted the seasonal autoregressive integrated moving average (SARIMA), which can capture the seasonal feature of time series, to predict the air quality in Kerala. However, due to the complexity and uncertainty of air quality prediction tasks, it is difficult for the statistical methods to perform well for long-term predictions. Different from the statistical methods, machine learning methods are non-parametric methods that can automatically learn new patterns, and thus can handle the complex non-linearity of temporal data. In recent years, machine learning methods have been widely employed for air quality prediction, including the support vector regression (SVR) \cite{liu2017urban}, the extreme gradient boosting (XGBoost) \cite{pan2018application} algorithm, and the random forest approach \cite{yu2016raq}, etc. However, these methods do not take into account the Spatio-temporal correlations and thus limiting their prediction performance.

To extract the Spatio-temporal correlations, deep learning methods have been applied for air quality forecasting. Wen et al. \cite{wen2019novel} proposed a spatiotemporal convolutional long short-term memory (LSTM) neural network to capture the temporal and spatial dependencies. The temporal patterns were captured through the LSTM networks and the spatial dependencies were extracted by the convolutional neural networks (CNNs). Zhang et al. \cite{zhang2020pm2} then modeled the Spatio-temporal correlations with the CNN model and the gated recurrent units (GRUs). The above methods can provide satisfactory prediction results, nevertheless, the CNN model is not suitable to model the non-Euclidean structure data and thus the spatial relationships between air sensors cannot be effectively modeled.

Most recently, graph-based deep learning methods have gai-ned popularity since they can process the non-Euclidean structure data by modeling it to a graph for training. Wang et al. \cite{wang2021modeling} and Zhang et al. \cite{zhang2020pm2} separately employed graph convolutional networks (GCNs) to model the contextual relationships among air quality stations and further predict the air quality in the future. In a relatively short period, this modeling approach was very successful. 

Models based on GCN need to construct the graph structure in advance. Traditional methods for constructing graph structures are usually based on prior knowledge, which can be divided into three categories: methods based on geographic distance, time-series similarity \cite{wang2021modeling}, and wind field information \cite{wang2020pm2}. However, we cannot exhaustively enumerate all factors previously. Besides, parallel learning of too many graphs may result in too many parameters and high computational costs. In conclusion, inaccurate prior information may lead us to incorrectly connect two unrelated stations or lose links between two related stations. Moreover, the contextual relationships are constantly changing due to the impacts of the wind fields and other factors. Therefore, the dynamic graph is more suitable to model the relationships among stations in the real world. To overcome these limitations, we develop a new method that learns the dynamic links between two stations automatically.

In this paper, we propose to construct a Dynamic Graph Neural Network with Adaptive Edge Attributes (DGN-AEA). Firstly, to address the shortcomings of prior information, we propose a method that uses self-use dynamic graph learning. However, the dynamic adjacency matrix represents the connection relationships between nodes will change with time, which will bring instability and difficulty to the training of the model. So, we divide the adjacency matrix into two parts, the connection relation (topology) matrix, and the weight matrix, and propose to use an adaptive edge attributes (weights) matrix. Experiments show that the adaptive edge attributes can improve the prediction result. Secondly, in order to solve the physical consistency problem of many existing deep learning models, we designed a dynamic edge connection construction method using wind field information and combined it with adaptive edge connection through the method of multi-graph stitching. Through this way, these learnable edges can be used as a correction of prior information, which can help the model to get rid of the one-sidedness of prior knowledge. Thirdly, we also calculate the outbound and inbound directions respectively when aggregating the neighbor node’s information. Through this way, the inflow and outflow processes during the diffusion of pollutants are simulated. In summary, the contributions of this paper are listed as follows.

\begin{itemize}
	\item We introduce the adaptive dynamic graph learning unit to learn the dual-path weighted edges automatically, to solve the problem of correlation graph modeling in non-Euclidean space.
    \item The wind field data can be integrated into our model as a type of directed dynamic connection by a Multi-Graph Process Block (MGP). The physical consistency of the model is improved in this way.
	\item For each node, we calculate its in-degree and out-degree separately to model convolution calculations on weighted directed graphs, which is more suitable for complex systems in the real world.
    \item The proposed DGN-AEA model improves the prediction capabilities and achieves state-of-the-art prediction accuracy.

\end{itemize}

The remainder of this paper is organized as follows. In Section~\ref{Method}, we introduce the method to construct the adaptive dynamic graph, and our proposed DGN-AEA model. In Section~\ref{Experiment design}, we describe the data used in our research and how we design experiments to verify the performance of DGN-AEA on the real-world air quality dataset. In Section~\ref{Results}, we show results of experiments and try to discuss what makes DGN-AEA performs better. Finally, we conclude and discuss the future works in Section~\ref{Conclusions}.

% \section{Materials and methods}
% \label{Materials and methods}

\section{Method used}
\label{Method}

In this section, we firstly give the mathematical definition of the air quality prediction. Next, we describe how we construct the two kinds of dynamic graphs. Then, as illustrated in Figure~\ref{framework}, we introduce the DGN-AEA model which is designed to solve the adaptive graph learning problem. We show the details of how we leverage the framework of GCNs on the spatial domain to handle message passing on directed edges. We also use the spatial block Dynamic Multi-Graph Process Block(MGP) to combine the adapted edge-attributes and the wind graph with MLPs, and the temporal block GRU. Finally, we form the stacked GCNs, which need spatial and temporal blocks working together to capture the spatio-temporal dependencies among cities.

\begin{figure}[ht!]
  \centering
    \includegraphics[width=8cm]{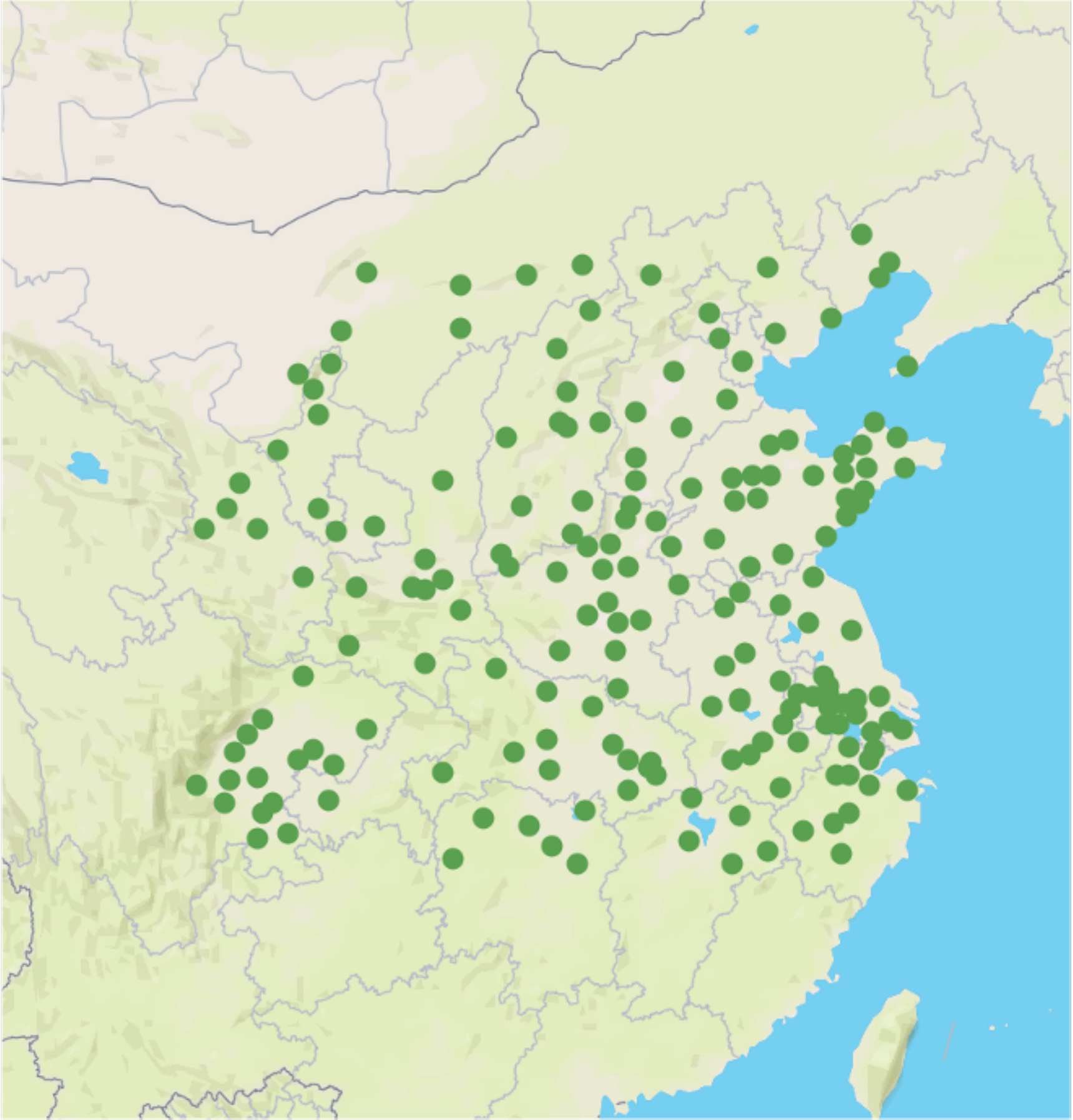}\\
    \vspace{0.0cm}
    \caption{The distribution of the observation station of the pollutant studied in our work. Each point represents one city-level observation station.}
    \label{stations}
\end{figure}

\subsection{Problem definition}
Air quality prediction can be seen as a typical spatial-temporal prediction problem. Let $X^t\in R^N$ denote the observed $\mathrm{PM}_{2.5}$ concentrations at time step $t$. The method based on GCNs usually models the changing spatial correlations among different cities by the dynamic directed graph $G=(V, E, t)$, where $V$ is the set of nodes and it is always the number of $N$. $E^t$ is the set of weighted edges representing the potential interactions among cities where its weight may change over time. Let $S^t\in R^{N\times s}$ denote the nodes’ attribute and $Z^t\in R^{N\times z}$ denote the edges’ attribute at time step $t$, where $s$ and $z$ represent the variable dimensions of node features and edge features, respectively. The problem aims to predict the next $T$ steps of $\mathrm{PM}_{2.5}$ concentrations $[X^{t+1}, ... , X^{t+T}]$. based on the nodes’ attribute $[S^{t+1}, ... , S^{t+T}]$ and the edges’ attributes $[Z^{t+1}, ... , Z^{t+T}]$. The mapping among the input and output can be shown as follows:

\begin{equation}
\left[X^t;S^{t+1},\ \cdots,\ S^{t+T};Z^{t+1},\cdots,Z^{t+T}\right]\stackrel{f(\cdot)}{\longrightarrow}\left[{\hat{X}}^{t+1},\ \cdots,\ {\hat{X}}^{t+T}\right]
\label{eq1}
\end{equation}

where ${\hat{X}}^t$ represents the predicted vector, and $f(\cdot)$ is the prediction function based on the DGN-AEA framework.

\begin{figure*}[ht!]
  \centering
    \includegraphics[width=18.5cm]{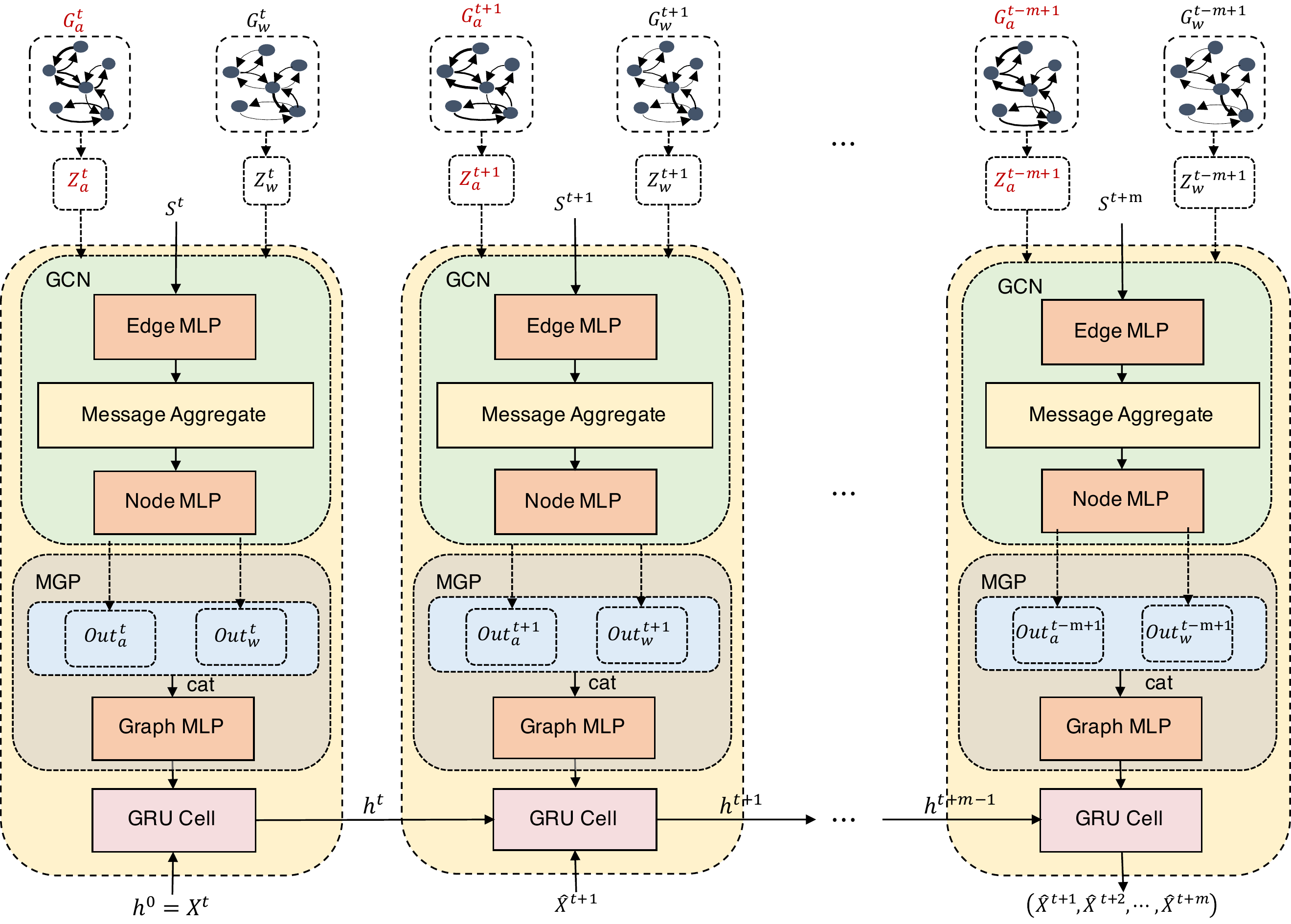}\\
    \vspace{0.0cm}
    \caption{Model structure of proposed model DGN-AEA.}
    \label{framework}
\end{figure*}

\subsection{Dynamic graph construction}
\label{Graph Construction}

In the air quality forecasting problem, we need to predict the future steps of all the cities. So the number of nodes will not change with time, which is different from some evolving dynamic graph problems \cite{barabasi1999emergence, leskovec2005graphs, you2018graphrnn}.

We dfine the weighted adjacency matrix $A$, which it can be divided into two parts: the topology matrix ($P$) of 0 or 1 indicating whether two nodes are connected, and the weight matrix (represents the edge attributes $Z$) indicating the strength of mutual influence between nodes. When using a neural network for training, if the adjacency matrix changes with time, it will bring great instability during training. Therefore, we use the adaptive edge attribute to represent this changing node interaction since it does not change the connection relationship between nodes but changes the strength of these connections, which means that the topology matrix will be static but the adjacency matrix will be dynamic. Thus, we suppose that all the correlations among cites are decided by the Euclidean distance (Equation~\ref{eq4}) like many previous graph-based air quality prediction approaches.

\subsubsection{Topology and adjacency matrix}

As we all know, the pollutant (such as $\mathrm{PM}_{2.5}$, $\mathrm{PM}_{10}$) concentrations in one place are strongly affected by other adjacents. Considering that relationships like that in real world is usually sparse and different, to model these two spatial correlation features explicitly, we define the adjacency matrix $A$ of Graph $G$ as:

\begin{equation}
    A=P\odot Z,
\label{eq2}
\end{equation}

where $\odot$ represents the Hadamard product, $Z$ represents the edge attributes matrix. The formulation of $Z$ will show in the following sections. 

As to the topology metrix $P$, we introduce the effect of distance on site relevance for the impact of the site is inversely proportional to the distance. And when the altitude between the two stations is too high, the connections will also be blocked out. To consider the above two factors, we use a Heaviside step function to filter out the edges that do not meet the rules as follows:

\begin{equation}
P_{ij}=H\left(d-\xi_d\right)\cdot H\left(a-\xi_a\right),
\label{eq3}
\end{equation}

where $H(\cdot)$ is the Heaviside step function:

\begin{equation}
    H\left(x\right)=\\
    \begin{cases}
    1 ,& \text{x\textgreater0}\\
    0 ,& \text{otherwise}
    \end{cases},
\label{eq4}
\end{equation}

where $a$ represents the altitude of each node. $d$ is the Euclidean distance calculated by the relative positions between two stations:

\begin{equation}
     d=\sqrt{\left(x_1-x_2\right)^2+\left(y_1-y_2\right)^2}.
\label{eq5}
\end{equation}

Here the Euclidean distance and altitude threshold $\xi_d$ and $\xi_a$ equals 300km and 1.2km respectively. Here we get the topology matrix $P$.

\subsubsection{Node attributes}

The node attributes $S^t$ are mainly meteorological data. Which includes 17 types of variables same as \cite{wang2020pm2}. We chose 8 of them as the final node attributes: \emph{Temperature}, \emph{Planetary  Boundary Layer height}, \emph{K index}, \emph{Relative humidity}, \emph{Surface pressure}, \emph{Total precipitation} and \emph{the u and v component of wind}. The time interval is consistent of these node attributes with the $\mathrm{PM}_{2.5}$ concentration data as 3h.

\subsubsection{Edge attributes}

In our work, there are two kinds of edge attributes: One is from the wind field and another is from the adaptive neural network parameter. We use the advection coefficient as attributes from wind data and calculate it as follows:

\begin{equation}
    Z_{w}^t=relu\left(\frac{\left|{\vec{v}}^t\right|}{d}\cdot\cos{\left(\alpha-\beta\right)}\right),
\label{eq6}
\end{equation}

where ${\vec{v}}^t$ represents the wind speed at time $t$, $d$ is the distance between stations, and $\alpha$ and $\beta$ are angles of cities and wind directions. $relu(\cdot)$ is the ReLU activation function.

$Z_{a}^t\in R^{1\times l}$ the adaptive neural network parameters, where $l$ represents the number of edges, i.e., the number of 1 in the topology matrix $P$. We set it as one important parameter which can be seen as another kind of useful edge attribute in addition to wind effects in the air quality prediction problem. This parameter can be obtained by continuous iterative optimization through the training stage.

By setting adaptive dynamic edge weights as learnable parameters, such dynamic correlations can be directly learned during the end-to-end training process. Even in practical scenarios where some prior information is missing, the correlation network between sites can still be adaptively learned for spatio-temporal prediction. When using wind field information, we can consider this learnable parameter as a supplement to wind field information. The prediction accuracy can be further improved in this way. The details will be presented in Section~\ref{Experiments}.

\subsection{Dynamic Graph Neural Network with Adaptive Edge Attributes}

\subsubsection{Graph convolution block}

Many dynamic graph neural networks methods are based on the spectral domain. The convolution operation on the graph is equivalent to the product in the spectral domain after the Fourier transform. The corresponding Fourier transform basis is the eigenvector of the Laplace matrix. And the model Chebnet \cite{defferrard2016convolutional} uses Chebyshev polynomial to approximate the spectral convolution. However, these methods cannot handle the directed graph since the Laplacian matrices are used for undirected graphs \cite{tong2020directed,ma2019spectral}. They are not suitable for complex system modeling because many relations in complex systems are directed. Besides, the prediction accuracy is limited by the order of the Chebyshev polynomial fit, and in many cases does not perform as well as spatial GCNs \cite{skarding2021foundations}. To solve these problems, we take the spatial domain GCN i.e., the Message Passing Neural Network (MPNN) in use.

MPNN framework can be divided into two stages: message passing stage and readout stage \cite{gilmer2017neural}. Compared with spectral-domain GCN which can only model node attributes, MPNN directly aggregates messages from neighbor nodes and can also model edges, which makes it more flexible and intuitive. For node $i$ at time $t$, our GCN block with MPNN framework will work as the following equations:

\begin{equation}
    \varepsilon_i^t=\left[{\hat{X}}_i^{t-1},S_i^t\right],
\label{eq7}
\end{equation}

\begin{equation}
    m_{ij}^t=\varphi\left(\left[\varepsilon_i^t,\varepsilon_j^t,Z_{ij}^t\right]\right),
\label{eq8}
\end{equation}

\begin{equation}
    e_i^t=\omega\left(\sum_{j\in N\left(i\right)}\left(m_{ij}^t-m_{ji}^t\right)\right),
\label{eq9}
\end{equation}

where $[\cdot, \cdot]$ represent the concat operator that concats two 1-D vector to a single vector. Equation~\ref{eq7} represents the splicing operation of neighbor information and edge connection weights. $\varepsilon_i^t$ in Equation~\ref{eq7} represents the result after the concatenation operation of the input matrix. $\varphi(\cdot)$ in Equation ~\ref{eq8} represents one layer of MLP. $N(i)$ represents the neighbours of node $i$. The whole Equation~\ref{eq8} represents the process of aggregating neighbor information according to the edge weights. $m_{ij}$ and $m_{ji}$ respectively represent the in-degree and out-degree information of the node. In Equation~\ref{eq9}, $\omega\left(\cdot\right)$ is another layer of MLP. After Equation~\ref{eq9}, we can calculate the increase and decrease after the node’s message passing process.

In our proposed method, we use two kinds of edge attributes. For each different edge attribute, $Z$ may represent $Z_{w}$ or $Z_{a}$, then we can get $e_{w}$ and $e_{a}$ following Equation~\ref{eq9}. Next, we concat the two graph-level embeddings through the transfer layer:

\begin{equation}
    \zeta^t=\psi\left([e_{w}^t,e_{a}^t]\right).
\label{eq10}
\end{equation}

It should be noted that $Z_{a}$ is a learnable parameter in our model, thus it will be updated through the model training stage. After that, we can get an adaptive edge attribute $Z_{AEA}$.

Since the $\mathrm{PM}_{2.5}$ transport graph network is a directed graph, in order to realize the material conservation of source and sink nodes, we calculate the message aggregation process of each node's incoming and outgoing edges respectively. This is consistent with the physical process of pollutant diffusion. It can improve the prediction accuracy of the model. The specific calculation process is shown in the following Figure~\ref{fig3}.

\begin{figure}[ht!]
  \centering
    \includegraphics[width=8cm]{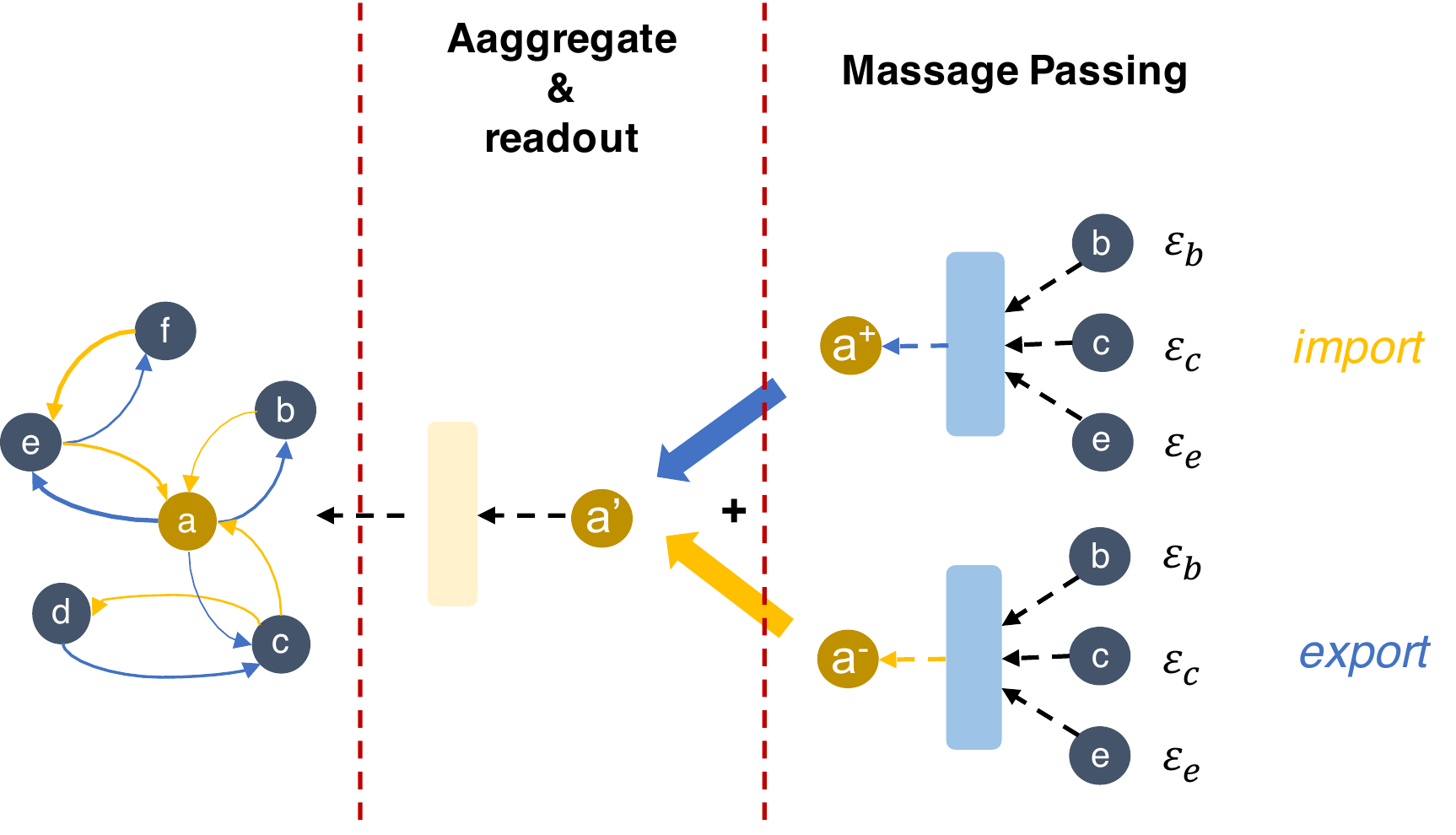}\\
    \vspace{0.0cm}
    \caption{Illustration for our used MPNN to process incoming and outgoing edge processing separately (use node a as example). Different colors of areows represents the import and export process separately.}
    \label{fig3}
\end{figure}

\subsubsection{Temporal processing block}

Our model DGN-AEA in Figure~\ref{framework} can be seen as the stacked structure: it is processed by MPNN, which is good at processing information on graphs, and then input the aggregated results to the GRU, which is good at iterative prediction of time series. Such a structure can more clearly extract effective information in the temporal and spatial domains, respectively. Here we will introduce the temporal processing block GRU. The input of GRU includes the historical $\mathrm{PM}_{2.5}$ concentrations of the nodes and future meteorological data, which is embedded in the node embedding $\varepsilon_i^t$. We also input the information gathered on the graph to the GRU. The prediction process for node $i$ at time $t$ is shown as follows:

\begin{equation}
    c_i^t=\left[\zeta_i^t,\varepsilon_i^t\right].
\label{eq11}
\end{equation}

This concatenates the input variables (results after graph convolution and the meteorological variables) to prepare for subsequent matrix operations:

\begin{equation}
    q_i^t=\sigma\left(W_q\cdot\left[h_i^{t-1},c_i^t\right]\right),
\label{eq12}
\end{equation}

\begin{equation}
    r_i^t=\sigma\left(W_r\cdot\left[h_i^{t-1},c_i^t\right]\right),
\label{eq13}
\end{equation}

where $q_i^t$ and $r_i^t$ in Equation~\ref{eq12} represent the results after operations of update gate and forget gate in GRU respectively. $\sigma(\cdot)$ represents the sigmoid activate function. $h_i^{t-1}$ is the hidden state of the previous time step, $c_i^t$ represents the aggregated input features after Equation~\ref{eq11}. The update gate is used to control the degree to which the state information of the previous moment is brought into the current state. The state information brought in at the previous moment is somewhat positively related to the value of the update gate. The reset gate controls how much information from the previous state is written to the current candidate set ${\widetilde{h}}_i^t$. The smaller the reset gate, the less information from the previous state is written. ${\widetilde{h}}_i^t$ is calculated by:

\begin{equation}
    {\widetilde{h}}_i^t=tanh\left(\widetilde{W}\cdot\left[r_i^t\ast h_i^{t-1},c_i^t\right]\right),
\label{eq14}
\end{equation}

where $W_q$, $W_r$, $\widetilde{W}$ in Equation~\ref{eq12} to Equation~\ref{eq14} are learnable parameters. After the gate control signal is obtained, we first use the reset gate to obtain the data after ``reset''. After the Hadamard product operation of the reset gate $r_i^t$ and the hidden layer information of the previous step $h_i^{t-1}$, then spliced the input signal $c_i^t$ of the current step. Then we multiply the result of the above operation by a learnable matrix, and then let the result pass a \emph{tanh} activation function to scale the result to the range of [-1, 1]. As shown in Equation~\ref{eq14}. The input information is added to the current hidden state in a targeted manner, which is equivalent to memorizing the state at the current moment.

Finally, Equation~\ref{eq15} performs the memory and updates operations at the same time, and obtains the updated hidden layer state:

\begin{equation}
    h_i^t=\left(1-q_i^t\right)\ast h_i^{t-1}+q_i^t\ast{\widetilde{h}}_i^t,
\label{eq15}
\end{equation}

After the whole operation, we finally get the prediction result of $\mathrm{PM}_{2.5}$ concentration by:

\begin{equation}
    {\hat{X}}_i^t=\Omega\left(h_i^t\right),
\label{eq16}
\end{equation}

where $\Omega\left(\cdot\right)$ is a MLP layer.

\subsubsection{Proposed learning algorithm}

We use the stacked spatiotemporal prediction structure. As shown in Figure~\ref{framework}, we first use MPNN to perform a convolution operation on the $\mathrm{PM}_{2.5}$ concentration of each site on the graph according to the edge weight. Two MPNN are used to process the wind field information map and adaptive dynamic map respectively. Then the two graph processing results are aggregated through a layer of MLP (as shown in Equation~\ref{eq7} to Equation~\ref{eq10}). Then, the output of the entire graph convolution part and the historical meteorological data are input into the GRU for time series iterative processing as Equation~\ref{eq11} to Equation~\ref{eq15}. Finally, we get the future forecast result as Equation~\ref{eq15}. The whole process of the proposed DGN-AEA is shown in Algorithm~\ref{alg}.

\begin{algorithm}[t]
  \renewcommand{\algorithmicrequire}{\textbf{Input:}}
  \renewcommand{\algorithmicensure}{\textbf{Output:}}
      \caption{$\mathrm{PM}_{2.5}$ Prediction Algorithm}
      \label{alg}
      \begin{algorithmic}[1]
      \REQUIRE{
          Historical $\mathrm{PM}_{2.5}$ concentrations $X^0$; \\
          Node’s attributes $S=[S_1, \cdots, S_T]$; \\
          Edge’s attributes(by wind) $Z_{w}=[Z_{w}^1,\cdots,\ Z_{w}^T]$;\\
          Randomly initialized adaptive edge’s attributes $Z_{a}=[Z_{a}^1,\cdots,\ Z_{a}^T$];
      }
      \ENSURE{
          Future PM2.5 concentrations $\hat{X}=[{\hat{X}}^1,\ ...,\ {\hat{X}}^{1+T}]$; \\
          Learned adaptive edge attributes $Z_{AEA}=[Z_{AEA}^1,\ ...,\ Z_{AEA}^{1+T}]$; \\
          Evaluation metric MAE and RMSE.
      }
      \STATE{
          ${\hat{X}}^0=X^0$\\
      }
      \STATE{
          $h^0=0$
      }
      \FOR{each time step $t \in [1,T]$}
          \FORALL{$v_i \in \mathbf{V}$}
              \STATE $\zeta_i^t=\psi\left(MPNN1\left({\hat{X}}_i^0,S_i^t,Z_{w}^t\right),MPNN2\left({\hat{X}}_i^0,S_i^t,Z_{a}^t\right)\right)$;
              \STATE ${\hat{X}}_i^t=\ GRU(\varepsilon_i^t,\zeta_i^t,h_i^{t-1})$;
              \STATE $\hat{X}=[\hat{X},{\hat{X}}_i^t]$.
          \ENDFOR
      \ENDFOR
      \STATE Calculate MAE and RMSE follow the Equation~\ref{eq18} and Equation~\ref{eq19};
      \RETURN $\hat{X}$, $Z_{AEA}$, MAE, RMSE
      \end{algorithmic}
  \end{algorithm}

\section{Data and experiment design}
\label{Experiment design}

In this section, we will show the details of our selected dataset and experiment settings. 

\subsection{Experiment settings}
Our experiments are conducted on Linux system with CPU: Intel(R) Xeon(R) Gold 5218 CPU @ 2.30GHz and GPU: NVIDIA Corporation Device 2204 (rev a1) The batch size of model training, validation, and test data are all 32. All models are trained up to 50 epochs by early stop rules with 10 steps and use RMSProp optimizer. The learning rate is 5e-4 and weight decay is also set to 5e-4. All the prediction results are the average result after 10 repetitions. In the training stage, we aim to minimize the Mean Square Error (MSE) Loss function as the following equation:

\begin{equation}
    MSE\ Loss=\ \frac{1}{T}\sum_{t=1}^{T}\left(\frac{1}{N}\sum_{i=1}^{N}\left({\hat{X}}_i^t-X_i^t\right)^2\right),
\label{eq17}
\end{equation}

\noindent where $T$ is the length of the prediction time step, and $N$ represents the number of samples. $\hat{X}$ and $X$ represent the predicted value and the ground truth of $\mathrm{PM}_{2.5}$ concentrations respectively.

To evaluate the prediction accuracy between models, we adopt two evaluation metrics: Mean Absolute Error (MAE) and Root Mean Square Error (RMSE).

\begin{equation}
    MAE=\frac{1}{N}\times\sum_{i=1}^{N}\left|{\hat{y}}_i-y_i\right|,
\label{eq18}
\end{equation}

\begin{equation}
    RMSE=\sqrt{\frac{1}{N}\times\sum_{i=1}^{N}\left({\hat{y}}_i-y_i\right)^2},    
\label{eq19}
\end{equation}

where $y$ is ground truth and $\hat{y}$ represents the prediction results given by models. These two are commonly used indicators to evaluate the accuracy of time series forecasting.

\subsection{Data used}
To examine the ability of the model to solve real problems, we conduct experiments on the real-world datasets from the previous work\cite{yi2018deep}. This dataset is collected from MEE\footnote[1]{\url{https://english.mee.gov.cn/}} and EAR5\footnote[2]{\url{https://climate.copernicus.eu/climate- reanalysis}}. It contains three types of data: Sites' geographic information, meteorological data, and pollutant concentration ($\mathrm{PM}_{2.5}$) data. The last two types of variables are time series data ranging from 2015-1-1 00:00:00 to 2018-12-31 23:59:59, with 3 hours for each time step. The dataset includes 184 city-level observation stations as shown in previous Figure ~\ref{stations}.

To test the predictive ability of the model under different circumstances. We divide the dataset into three parts by time. The training set of the first dataset is the data for two years in 2015 and 2016, and the test set and validation set are the data for the whole year of 2017 and 2018, respectively. The training, validation, and testing of data set 2 were sequentially intercepted in the winter of 2015 - 2018 for three consecutive years (November 1st - February 28th of the following year). This is because winter is usually the season of high $\mathrm{PM}_{2.5}$ pollution in China, and the average value of the data is higher. Dataset 3 uses the 2016 autumn and winter 4 months (September 1st - December 31st) for training, uses the winter data of the following two months for verification (December 1st to December 31st) and test (January 1st to January 31st of the following year). The period of 2016 was chosen because that winter saw almost the worst pollution in Chinese history.

\subsection{Baselines}
In our work, we consider baselines to examine the model effect. Baselines include classical statistical models, classical spatio-temporal prediction models, and state-of-the-art deep learning models with adaptive graph components.

\begin{table*}[ht]
    \centering
    \caption{Prediction accuracy compared with baselines.}
    \begin{tabular}{c|c|c|ccccc}
    \toprule
    \multirow{2}[2]{*}{Dataset} & \multicolumn{2}{c|}{Methods} & \multirow{2}[2]{*}{HA} & \multirow{2}[2]{*}{LSTM} & \multirow{2}[2]{*}{GC-LSTM} & \multirow{2}[2]{*}{$\mathrm{PM}_{2.5}$-GNN} & % \multirow{2}[2]{*}{\makecell[c]{Graph-WaveNet \\ (w/o weather)}} & 
    \multirow{2}[2]{*}{DGN-AEA} \\
          & \multicolumn{2}{c|}{Metric} &       &       &       &       & \\
    \midrule
    \multirow{8}[4]{*}{\begin{sideways}Dataset 1\end{sideways}} & \multirow{4}[2]{*}{RMSE} & 3    & \multirow{4}[2]{*}{25.81} & 12.17$\pm$0.10 & 12.03$\pm$0.07 & 11.55$\pm$0.11 & \textbf{11.34$\pm$0.07} \\
          &       & 6   &       & 15.61$\pm$0.11 & 15.40$\pm$0.07 & 14.76$\pm$0.10 &  \textbf{14.53$\pm$0.09} \\
          &       & 12   &       & 18.56$\pm$0.11 & 18.38$\pm$0.09 & 17.70$\pm$0.15  & \textbf{17.06$\pm$0.11} \\
          &       & 24   &       & 20.87$\pm$0.16 &  20.81$\pm$0.09 & 20.18$\pm$0.17  & \textbf{19.20$\pm$0.15} \\
\cmidrule{2-8}          & \multirow{4}[2]{*}{MAE} & 3    & \multirow{4}[2]{*}{37.26} & 9.43$\pm$0.09 & 9.31$\pm$0.06 & 8.93$\pm$0.05 & \textbf{8.75$\pm$0.05} \\
          &       & 6   &       & 12.44$\pm$0.12 & 12.25$\pm$0.06 & 11.70$\pm$0.10 & \textbf{11.50$\pm$0.08} \\
          &       & 12   &       & 14.88$\pm$0.14 & 14.69$\pm$0.10 & 14.11$\pm$0.17 & \textbf{13.56$\pm$0.11} \\
          &       & 24   &       & 16.48$\pm$0.15 & 16.45$\pm$0.08 &  15.90$\pm$0.19 & \textbf{15.06$\pm$0.14} \\
    \midrule
    \multirow{8}[4]{*}{\begin{sideways}Dataset 2\end{sideways}} & \multirow{4}[2]{*}{RMSE} & 3    & \multirow{4}[2]{*}{52.21} & 18.01$\pm$0.17 & 18.30$\pm$0.11 & 17.61$\pm$0.17 & \textbf{17.15$\pm$0.07} \\
          &       & 6   &       & 23.55$\pm$0.22 & 23.65$\pm$0.19 & 22.94$\pm$0.24 & \textbf{22.29$\pm$0.09} \\
          &       & 12   &       & 28.60$\pm$0.23 & 28.547$\pm$0.17 & 27.52$\pm$0.30 & \textbf{26.85$\pm$0.11} \\
          &       & 24   &       & 32.82$\pm$0.27 & 33.03$\pm$0.33 & 31.70$\pm$0.29 & \textbf{30.79$\pm$0.20} \\
\cmidrule{2-8}          & \multirow{4}[2]{*}{MAE} & 3    & \multirow{4}[2]{*}{35.84} & 14.01$\pm$0.15 & 14.21$\pm$0.08 & 13.70$\pm$0.15 & \textbf{13.33$\pm$0.06} \\
          &       & 6   &       & 18.90$\pm$0.19 & 18.95$\pm$0.19 & 18.38$\pm$0.23 & \textbf{17.83$\pm$0.09} \\
          &       & 12   &       & 23.14$\pm$0.21 & 22.96$\pm$0.16 & 25.26$\pm$0.41 & \textbf{21.60$\pm$0.10} \\
          &       & 24   &       & 26.61$\pm$0.32 & 26.37$\pm$0.36 & 25.26$\pm$0.41 & \textbf{24.45$\pm$0.22} \\
    \midrule
    \multirow{8}[4]{*}{\begin{sideways}Dataset 3\end{sideways}} & \multirow{4}[2]{*}{RMSE} & 3    & \multirow{4}[2]{*}{42.33} & 26.43$\pm$0.36 & 26.56$\pm$0.20 & 25.51$\pm$0.32 & \textbf{24.75$\pm$0.17} \\
          &       & 6   &       & 33.87$\pm$0.41 & 34.06$\pm$0.27 & 32.95$\pm$0.33 & \textbf{31.98$\pm$0.36} \\
          &       & 12   &       & 40.98$\pm$0.55 & 40.84$\pm$0.66 & 39.76$\pm$0.71 & \textbf{38.78$\pm$0.27} \\
          &       & 24   &       & 45.08$\pm$1.02 & 44.86$\pm$0.70 & 45.04$\pm$0.88 & \textbf{42.18$\pm$0.84} \\
    \cmidrule{2-8}          & 
    \multirow{4}[2]{*}{MAE} & 3    & \multirow{4}[2]{*}{29.31} & 20.52$\pm$0.28 & 20.63$\pm$0.18 & 19.84$\pm$0.27 & \textbf{19.22$\pm$0.14} \\
          &       & 6   &       & 27.14$\pm$0.40 & 27.23$\pm$0.23 & 26.37$\pm$0.32 & \textbf{25.53$\pm$0.34} \\
          &       & 12   &       & 33.16$\pm$0.59 & 32.92$\pm$0.58 & 32.14$\pm$0.74 & \textbf{31.19$\pm$0.24} \\
          &       & 24   &       & 36.89$\pm$1.01 & 36.62$\pm$0.73 & 36.23$\pm$0.99 & \textbf{34.08$\pm$0.78} \\
    \bottomrule
  \end{tabular}%
  \label{exp results}%
\end{table*}%

\begin{itemize}
\item \textbf{HA:} The Historical Average (HA) model is a typical time series analysis model, which main idea is to use the average of all the values at the corresponding time in history (known data) as the predicted value for future. Therefore, there is no concept of prediction time step. Here we refer to the construction method in the article \cite{li2017diffusion} to calculate the test set, and intercept all the moments of one week to predict the corresponding time points.
\item \textbf{LSTM:} The long short-term memory (LSTM) \cite{du2019deep} model is an improvement of the RNN model, which uses three types of gates to extract more useful related historical data.
\item \textbf{GC-LSTM:} GC-LSTM \cite{qi2019hybrid} is a model which uses two spectral-based GCNs embedded into the long-short term memory model to extract spatio-temporal features from data. 
%The dimension of the hidden layer and output layer is 32 and 1, respectively. The Chebyshev filter size is 2.
\item \textbf{$\mathrm{PM}_{2.5}$-GNN:} $\mathrm{PM}_{2.5}$-GNN \cite{wang2020pm2} is a state-of-the-art prediction model for $\mathrm{PM}_{2.5}$ concentrations prediction. It also uses the stacked spatio-temporal structure based on GCNs and RNNs.
\item \textbf{Graph WaveNet (w/o weather):} The Graph WaveNet  \cite{wu2019graph} develops a novel adaptive dependency matrix, which can automatically capture the spatial dependency from data. It uses Temporal Convolutional Network (TCN) as the temporal block. It has achieved state-of-the-art results in many real-world datasets, especially in traffic flow forecasts. Since the original model uses one-dimensional convolution to operate on only one variable, we do not use multi-dimensional meteorological information when reproduce the model. Results will shown in Section~\ref{ablation}.
\end{itemize}

\begin{figure*}[h!]
    \centering
    \includegraphics[width=1\textwidth]{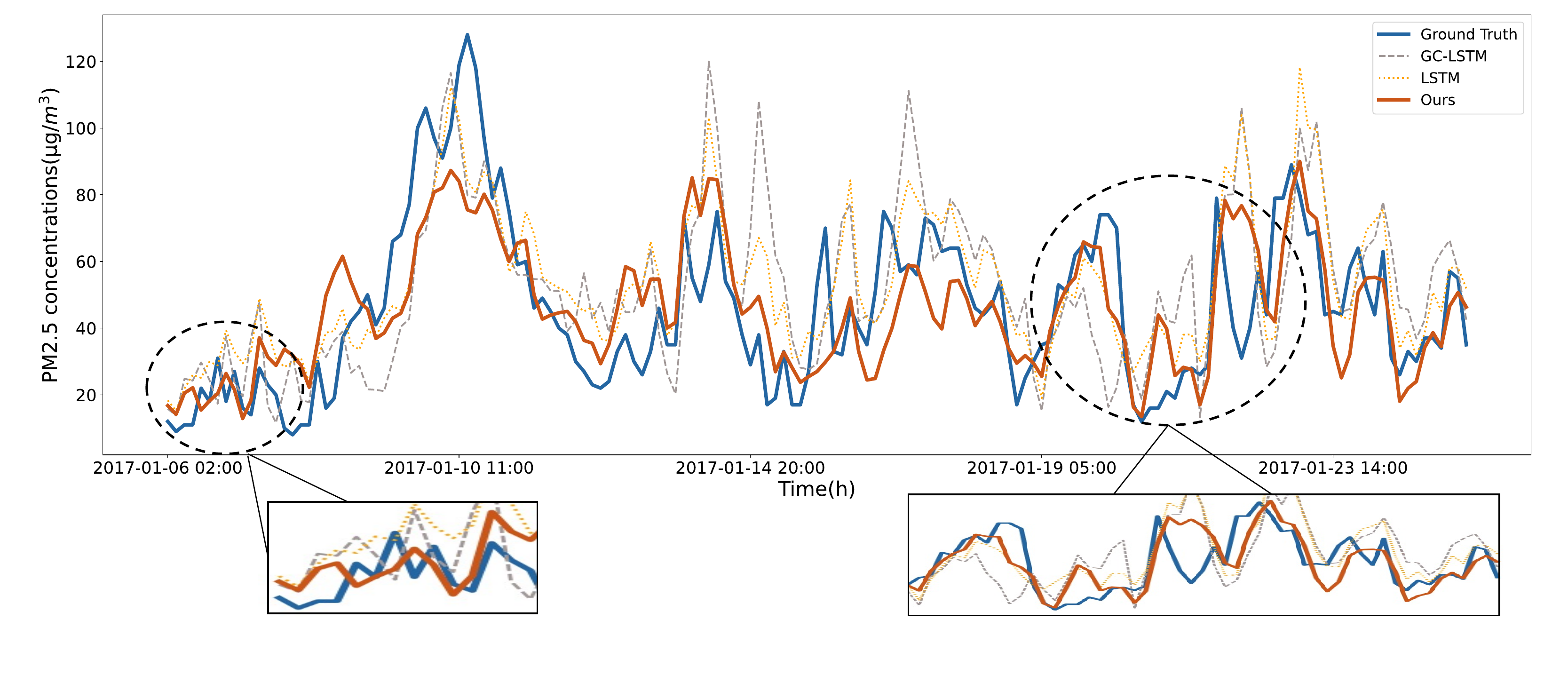}
    \caption{Future $\mathrm{PM}_{2.5}$ concentrations predicted by DGN-AEA.}
    \label{prediction}
\end{figure*}

\subsection{Ablation Study}

In order to further illustrate the role of the adaptive dynamic edge attribute, we compare the results with models with some parts removed, which are:

\begin{figure*}[h!]
    \centering
        \subfigure[ground truth]{
            \includegraphics[width=0.8\textwidth]{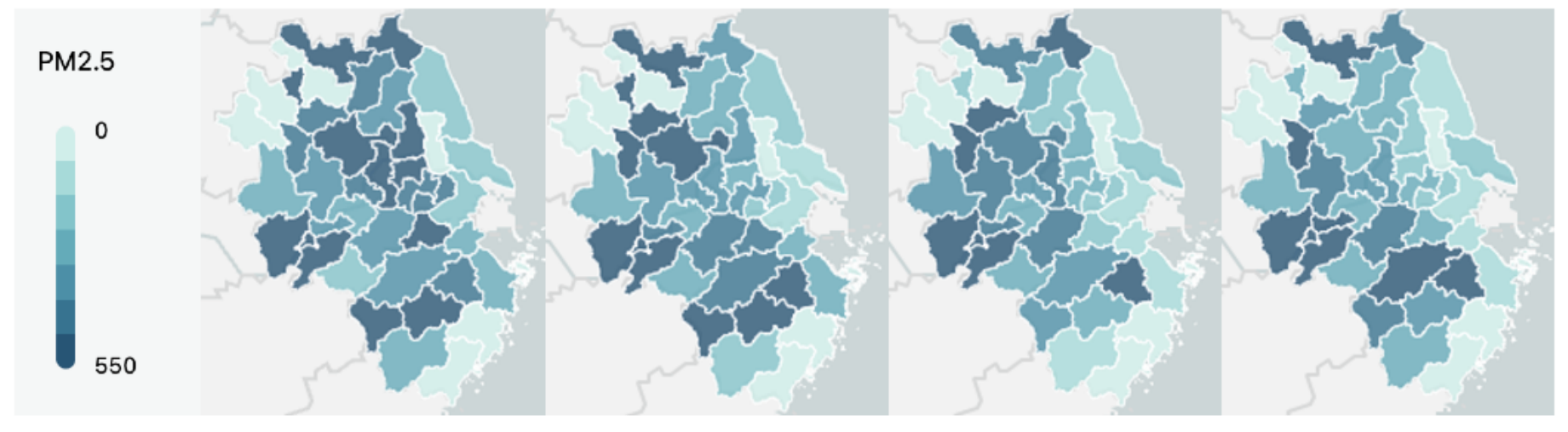}
            \label{csj-label}
        }
        \subfigure[predict result]{
            \includegraphics[width=0.8\textwidth]{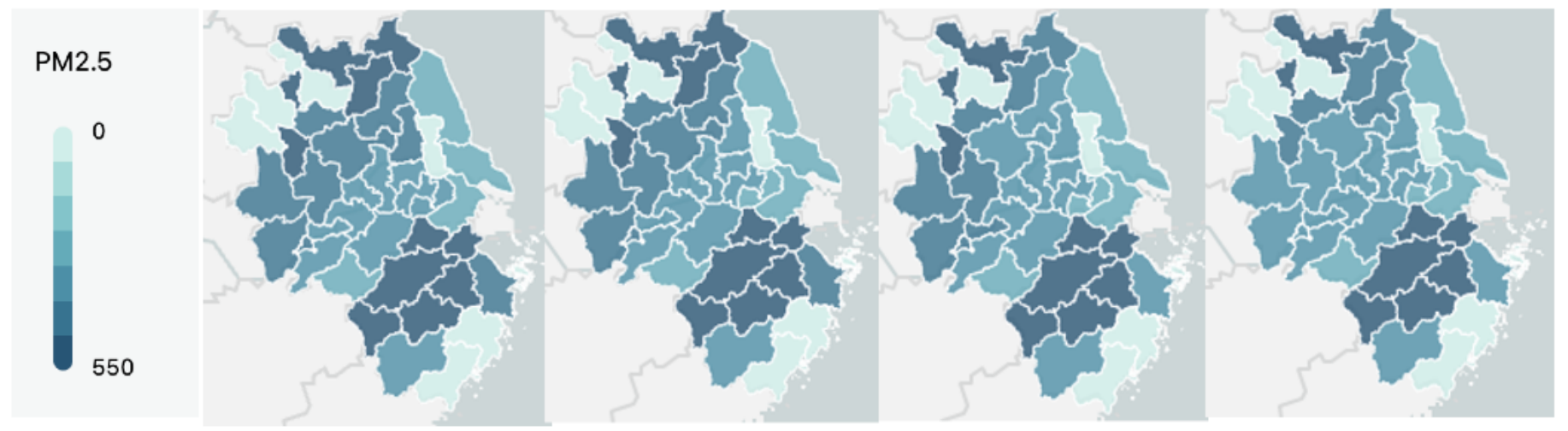}
            \label{csj-pred}
        }
         \subfigure[error between the two]{
            \includegraphics[width=0.8\textwidth]{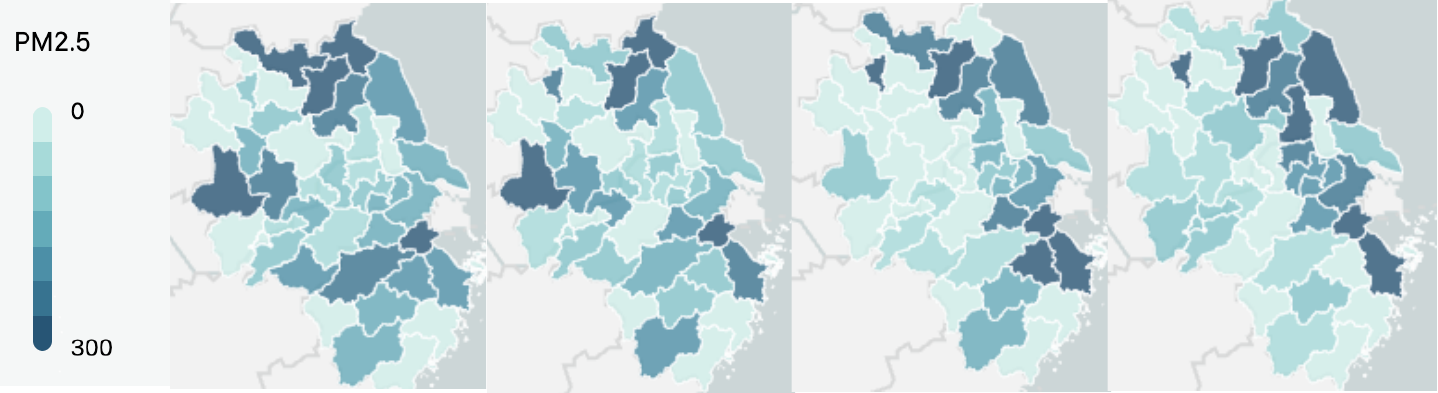}
            \label{csj-mae}
        }
    \caption{Visualization of forecast results in the Yangtze River Delta region. The data is derived from dataset 3 with 24 prediction horizon time steps. (The unit is $\mu g/m^3$)}
    \label{csj}
\end{figure*}

\begin{itemize}
\item \textbf{Static:} To demonstrate the role of dynamic graphs, we conduct experiments using only static graph structures based on distance and altitude calculations.
\item \textbf{Only AEA:} As described before, DGN-AEA integrates wind edge information. Here the wind information is removed and only adaptive edge attributes are used. This can illustrate the important role of wind in modeling $\mathrm{PM}_{2.5}$ forecasting.
\item \textbf{Only Wind:} Contrary to \textbf{Only AEA}, here we only use the wind field information and remove the adaptive edge attributes. Similar to the control variables approach, this can illustrate the importance of using adaptive edge attributes.
\item \textbf{W/O weather:} Here we also compare the effect of not inputting the future weather GRU module as known information to illustrate the effect of using future weather.
\item \textbf{AEA+Wind(ours):} As shown in Figure~\ref{framework}, we will use the multi-graph information of adaptive edge attributes and wind at the same time.
\end{itemize}

\section{Results and discussion}
\label{Results}

\subsection{Performance comparisions with baselines}

\begin{table*}[ht]
  \centering
  \caption{Result of ablation study.}
    \begin{tabular}{c|c|c|ccccc}
    \toprule
    \multirow{2}[2]{*}{Dataset} & \multicolumn{2}{c|}{Methods} & \multirow{2}[2]{*}{Only Wind} & \multirow{2}[2]{*}{Only AEA} & \multirow{2}[2]{*}{Static} & \multirow{2}[2]{*}{W/O weather} & \multirow{2}[2]{*}{AEA+Wind}  \\
          & \multicolumn{2}{c|}{Metric} &       &       &       & \\
    \midrule
    \multirow{8}[4]{*}{\begin{sideways}Dataset 1\end{sideways}} & \multirow{4}[2]{*}{RMSE} & 3 & 11.55$\pm$0.05 & 11.38$\pm$0.04 & 11.58$\pm$0.08 & 13.03$\pm$0.06 & \textbf{11.34$\pm$0.06} \\
          &       & 6   & 14.76$\pm$0.10 & 14.55$\pm$0.06 & 14.72$\pm$0.09 & 17.08$\pm$0.07 & \textbf{14.53$\pm$0.09} \\
          &       & 12   & 17.70$\pm$0.15 & 17.36$\pm$0.07 & 17.51$\pm$0.13 & 20.78$\pm$0.10 & \textbf{17.06$\pm$0.11} \\
          &       & 24   & 20.18$\pm$0.17 &  19.78$\pm$0.09 & 19.80$\pm$0.19 & 24.41$\pm$0.08 & \textbf{19.20$\pm$0.15} \\
    \cmidrule{2-8}          & 
    \multirow{4}[2]{*}{MAE} & 3 & 8.93$\pm$0.05 & 8.78$\pm$0.04 & 8.95$\pm$0.06 & 10.14$\pm$0.05 & \textbf{8.75$\pm$0.05} \\
          &       & 6   & 11.70$\pm$0.10 & 11.51$\pm$0.06 & 11.68$\pm$0.08 & 13.73$\pm$0.06 & \textbf{11.50$\pm$0.08} \\
          &       & 12   & 14.11$\pm$0.17 & 13.79$\pm$0.08 & 13.97$\pm$0.13 & 16.90$\pm$0.11 & \textbf{13.56$\pm$0.11} \\
          &       & 24   & 15.90$\pm$0.19 & 15.55$\pm$0.09 &  15.60$\pm$0.19 & 18.84$\pm$0.09 & \textbf{15.06$\pm$0.14} \\
    \midrule
    \multirow{8}[4]{*}{\begin{sideways}Dataset 2\end{sideways}} & \multirow{4}[2]{*}{RMSE} & 3 & 17.61$\pm$0.17 & 17.60$\pm$0.14 & 17.35$\pm$0.16 & 19.62$\pm$0.04 & \textbf{17.15$\pm$0.07} \\
          &       & 6    & 22.99$\pm$0.21 & 23.02$\pm$0.20 & 22.34$\pm$0.08 & 26.17$\pm$0.03 & \textbf{22.29$\pm$0.09} \\
          &       & 12   & 27.60$\pm$0.30 & 27.71$\pm$0.25 & 27.15$\pm$0.13 & 32.87$\pm$0.07 & \textbf{26.85$\pm$0.11} \\
          &       & 24   & 31.70$\pm$0.29 & 31.50$\pm$0.33 & 31.49$\pm$0.34 & 38.89$\pm$0.08 & \textbf{30.79$\pm$0.20} \\
    \cmidrule{2-8}          & 
    \multirow{4}[2]{*}{MAE} & 3 & 13.70$\pm$0.15 & 13.69$\pm$0.13 & 13.69$\pm$0.13 & 15.31$\pm$0.04 & \textbf{13.33$\pm$0.06} \\
          &       & 6   & 18.41$\pm$0.20 & 18.45$\pm$0.20 & 18.67$\pm$0.08 & 21.12$\pm$0.03 & \textbf{17.83$\pm$0.09} \\
          &       & 12   & 22.26$\pm$0.32 & 22.37$\pm$0.23 & 22.35$\pm$0.14 & 26.99$\pm$0.08 & \textbf{21.60$\pm$0.10} \\
          &       & 24   & 25.26$\pm$0.41 & 25.06$\pm$0.31 & 24.81$\pm$0.33 & 31.94$\pm$0.11 & \textbf{24.45$\pm$0.22} \\
    \midrule
    \multirow{8}[4]{*}{\begin{sideways}Dataset 3\end{sideways}} & \multirow{4}[2]{*}{RMSE} & 3 & 25.51$\pm$0.32 & 25.32$\pm$0.23 & 24.80$\pm$0.28 & 27.16$\pm$0.05 & \textbf{24.75$\pm$0.17} \\
          &       & 9   & 32.95$\pm$0.33 & 32.74$\pm$0.46 & 32.13$\pm$0.22 & 36.12$\pm$0.14 & \textbf{31.98$\pm$0.36} \\
          &       & 12   & 39.10$\pm$0.63 & 39.26$\pm$0.26 & 39.31$\pm$0.41 & 44.91$\pm$0.13 & \textbf{38.78$\pm$0.27} \\
          &       & 24   & 43.44$\pm$0.42 & 42.83$\pm$0.52 & 42.32$\pm$0.77 & 49.71$\pm$0.11 & \textbf{42.18$\pm$0.84} \\
    \cmidrule{2-8}          & 
    \multirow{4}[2]{*}{MAE} & 3   & 19.84$\pm$0.27 & 19.68$\pm$0.20 & 19.26$\pm$0.24 & 21.15$\pm$0.05 & \textbf{19.22$\pm$0.14} \\
          &       & 6   & 26.37$\pm$0.32 & 26.17$\pm$0.45 & 25.63$\pm$0.22 & 29.11$\pm$0.14 & \textbf{25.53$\pm$0.34} \\
          &       & 12   & 31.54$\pm$0.64 & 31.51$\pm$0.28 & 31.72$\pm$0.42 & 36.93$\pm$0.16 & \textbf{31.19$\pm$0.24} \\
          &       & 24   & 35.72$\pm$0.45 & 35.08$\pm$0.53 & 34.58$\pm$0.79 & 41.44$\pm$0.13 & \textbf{34.08$\pm$0.78} \\
    \bottomrule
  \end{tabular}%
  \label{ablation}%
\end{table*}%

We compare the prediction results with the evaluation metric between our DGN-AEA and baselines in all three datasets. In addition, we set different prediction horizon time steps with 3 (9h), 6 (18h), 12 (36h), and 24 (72h) so that we can compare the predictive ability of various models under different time prediction lengths. The best results are highlighted in bold-face in Table~\ref{exp results}.

We divided the data into three datasets. By designing the number of samples and seasons of the training set dataset, it can be considered that the training difficulty on the three datasets is increasing. The training set of data set 3 has the least data and the corresponding value is large, since winter is the high season of haze in China, the value of $\mathrm{PM}_{2.5}$ is generally higher.

It can be seen that our model always performs the best and the traditional statistical model HA is not always the worst. GC-LSTM performs a little better than LSTM, nevertheless, it does not perform well on our dataset overall. In dataset 3, our model DGN-AEA improves the RMSE of GC-LSTM by 6.81\%, 6.11\%, 5.04\%, and 6.7\%, respectively. $\mathrm{PM}_{2.5}$-GNN performs better than other baselines, but our model is more accurate. On the RMSE of dataset3, DGN-AEA is 2.98\%, 2.94\%, 2.46\%, 6.3\% more accurate than $\mathrm{PM}_{2.5}$-GNN. 

Compared with another adaptive graph model, Graph WaveNet requires a large number of parameters and has a high computational resource overhead, so the training is slow. Its effect is also not good. Compared with Graph WaveNet, our proposed model DGN-AEA improves the RMSE of dataset 3 by 47.46\%, 45.32\%, 44.56\%, 43.98\%, and 38.08\%, 36.05\%, 37.61\%, 36.99\% on MAE. We speculate that it may be due to the construction of the adjacency matrix that changes from time to time brings great difficulty to training on $\mathrm{PM}_{2.5}$ datasets, making it difficult for the model to grasp the exact topology of the stations.

The prediction fit curves of Linan are also plotted in Figure~\ref{prediction}.

The above results are the average of the whole map and fit curves for individual cities. In order to examine the prediction ability of the model at the local regional scale, we select the ground truths and predicted results of the Yangtze River Delta region for a continuous period of time to visualize. The result is shown in the figure~\ref{csj}.

\subsection{The function of dynamic graph}
To explain why we choose to retain two types of graph edges, we conduct ablation studies to explore the predictive effects of using different edges. We find that our proposed model is the best in most cases, regardless of the dataset or prediction at any time scale (Table~\ref{ablation}).

We can see that the models where we do not use dynamic information perform worse than the dynamic graph method in the vast majority of cases. 

In addition, the prediction accuracy of \emph{Only Wind} and \emph{Only AEA} are similar, and in most cases \emph{Only AEA} has even better MAE and RMSE metrics. Especially in Dataset 3, our proposed DGN-AEA has an average of 4.3\%,4.7\%, 3.4\%, 7.8\% MAE metric decrease than the other two dynamic edge attribute.

\begin{figure*}[h]  %figure 6
    \centering
    
    \subfigure[MAE on Dataset 1]{
     \begin{minipage}{0.32\textwidth}
    \includegraphics[width=1\textwidth]{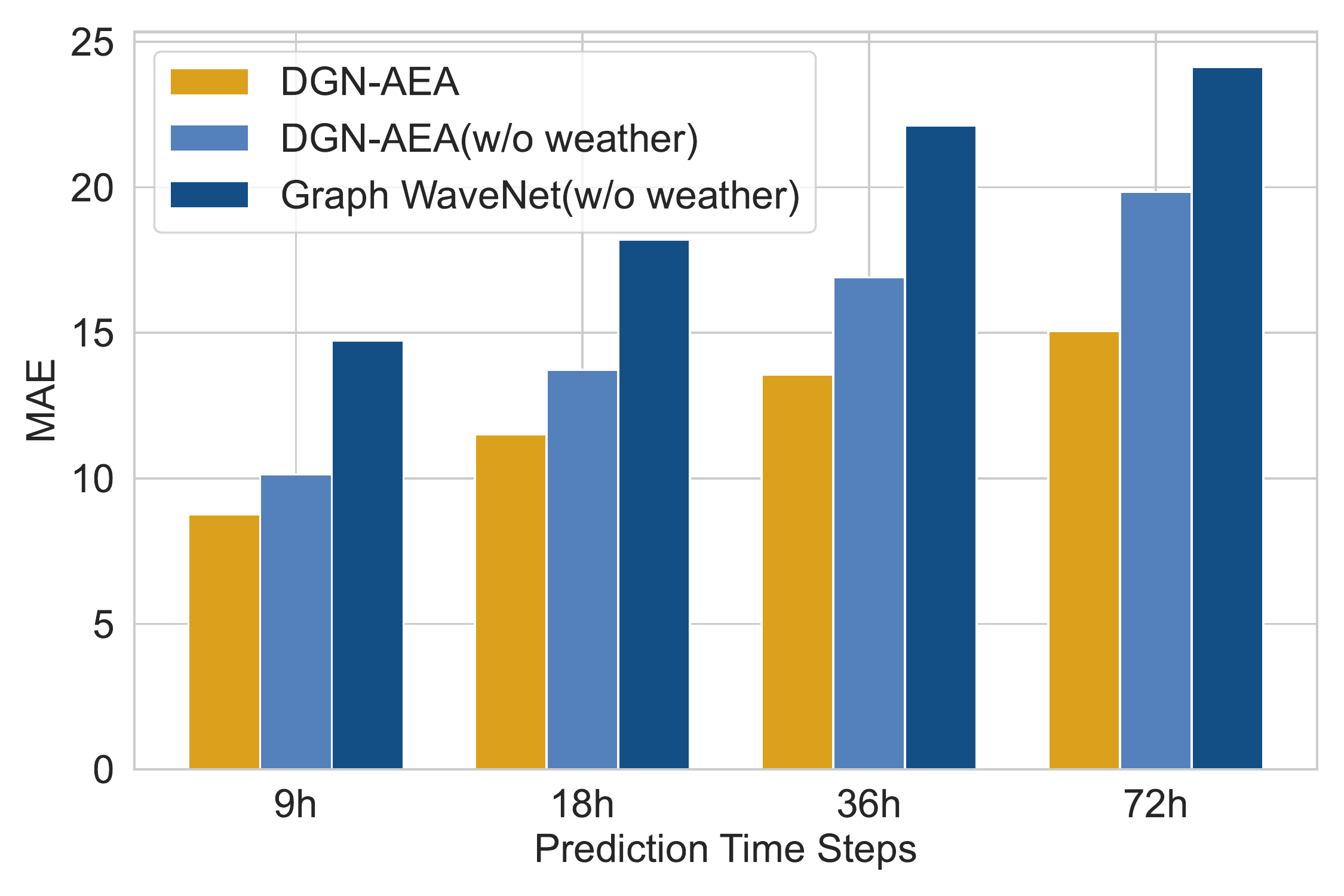}
    \label{bar1}
    \end{minipage}}
    \subfigure[MAE on Dataset 2]{
     \begin{minipage}{0.32\textwidth}
    \includegraphics[width=1\textwidth]{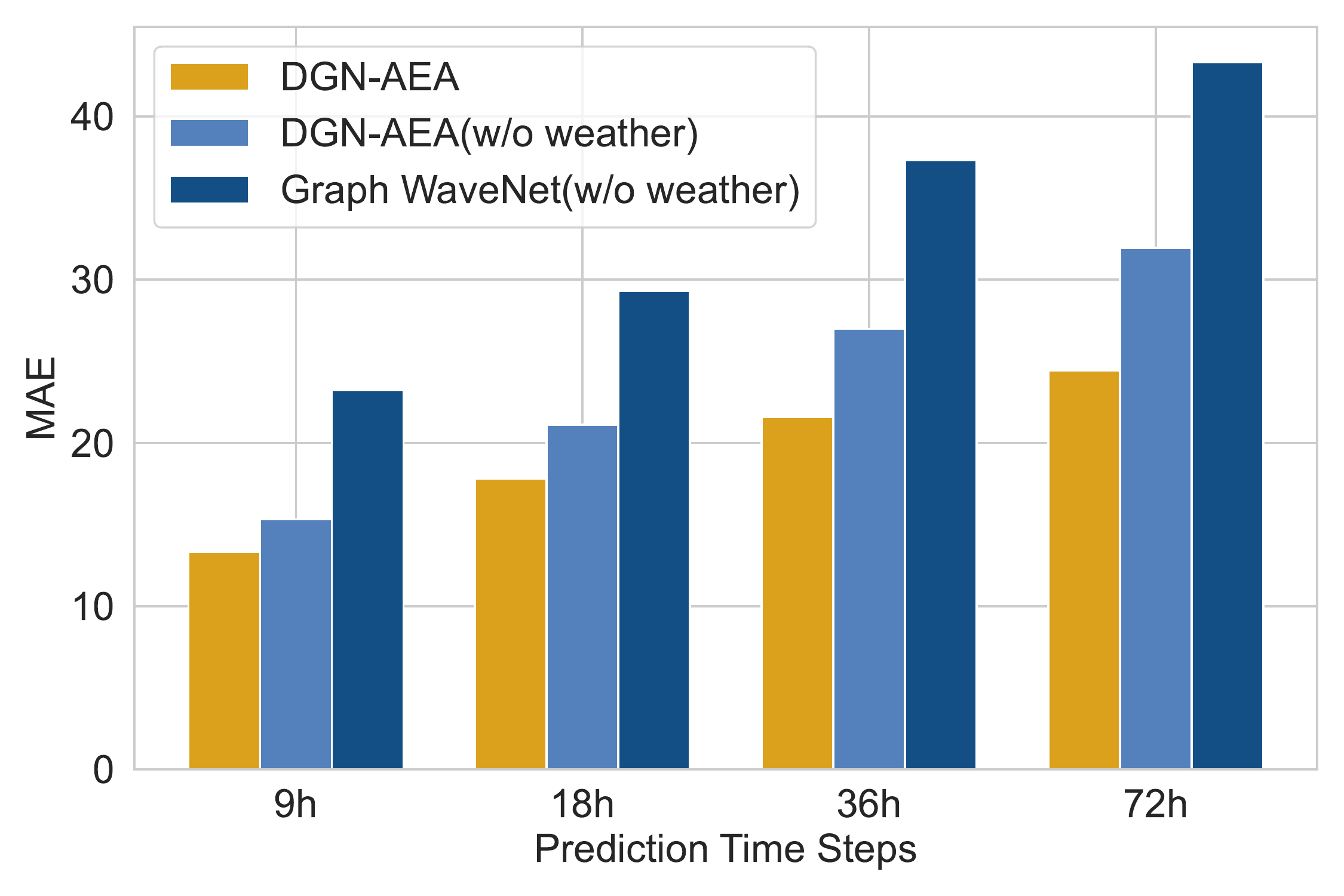}
    \label{bar3}
    \end{minipage}}
    \subfigure[MAE on Dataset 3]{
     \begin{minipage}{0.32\textwidth}
    \includegraphics[width=1\textwidth]{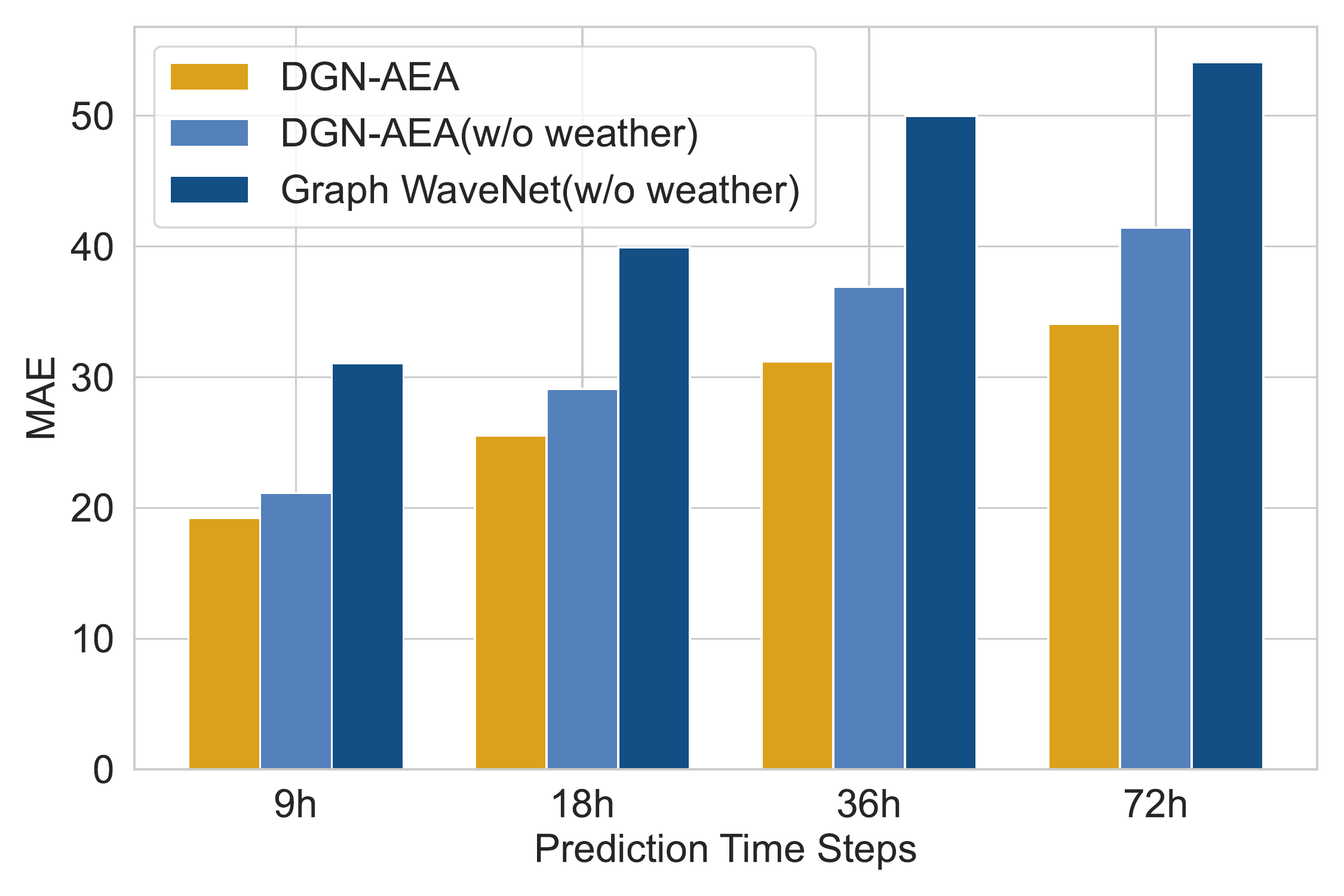}
    \label{bar5}
    \end{minipage}}
    \subfigure[RMSE on Dataset 1]{
     \begin{minipage}{0.32\textwidth}
    \includegraphics[width=1\textwidth]{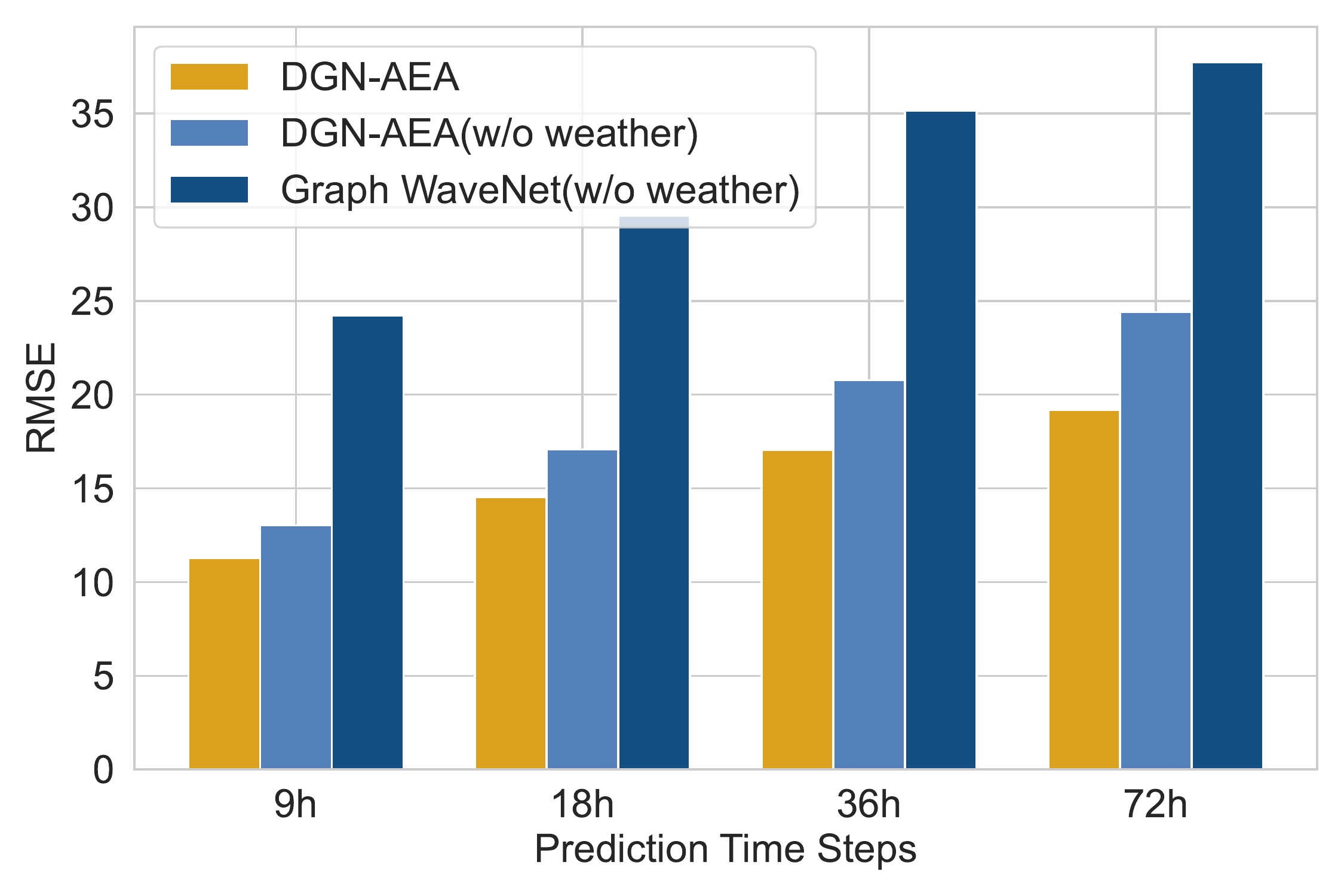}
    \label{bar2}
    \end{minipage}}
    \subfigure[RMSE on Dataset 2]{
     \begin{minipage}{0.32\textwidth}
    \includegraphics[width=1\textwidth]{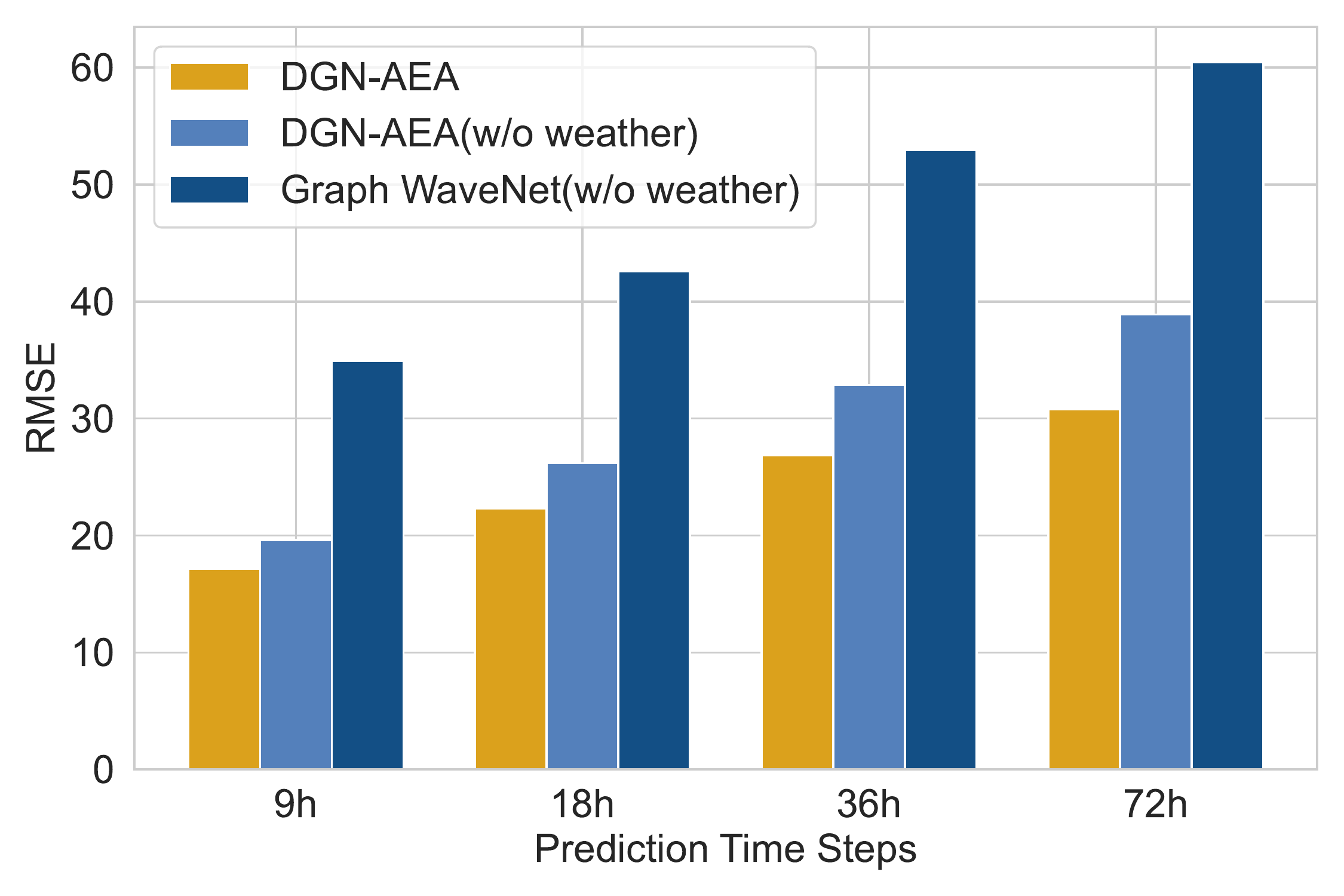}
    \label{bar4}
    \end{minipage}}
    \subfigure[RMSE on Dataset 3]{
     \begin{minipage}{0.32\textwidth}
    \includegraphics[width=1\textwidth]{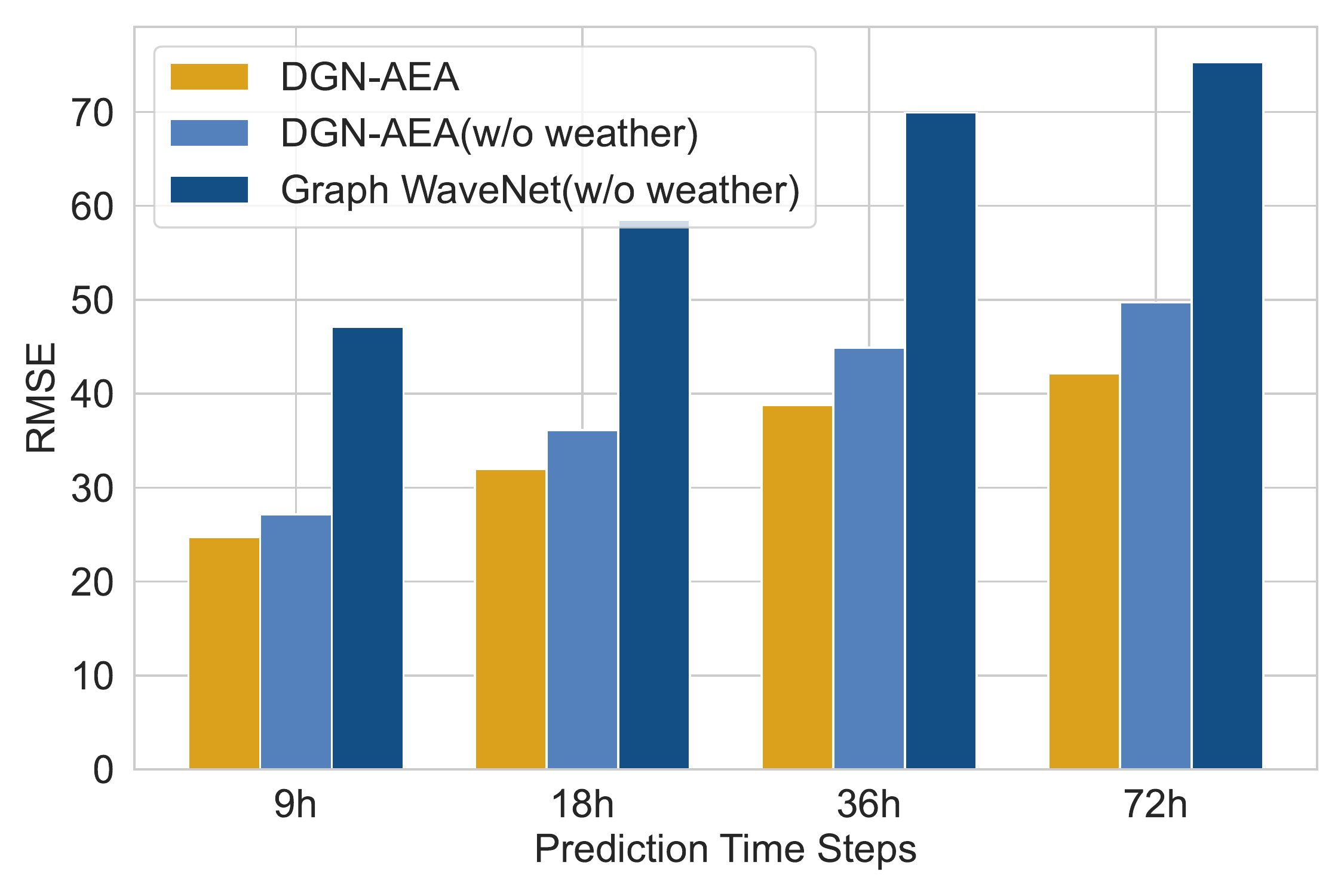}
    \label{bar6}
    \end{minipage}}

    \caption{Comparison among DGN-AEA, DGN-AEA (w/o weather) and Graph WaveNet(w/o weather) models. The results are the average of ten training sessions.}
\label{barplots}
\end{figure*}

\begin{figure*}[h!]
  \centering
   \subfigure[adaptive edge attributes]{
     \begin{minipage}{0.32\textwidth}
    \includegraphics[width=1\textwidth]{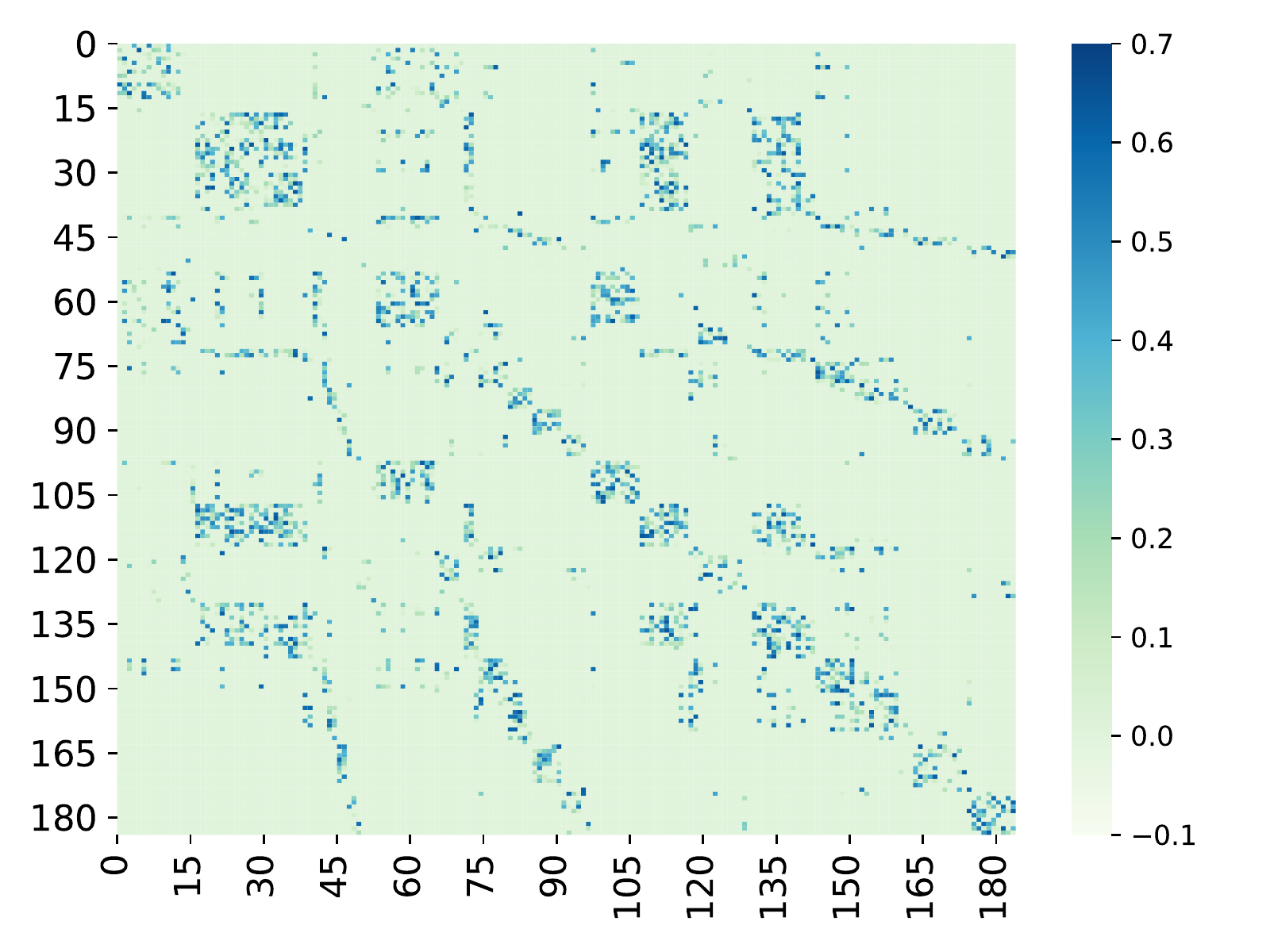}
    \label{pre3}
    \end{minipage}}
    \subfigure[edge attribute by wind]{
     \begin{minipage}{0.32\textwidth}
    \includegraphics[width=1\textwidth]{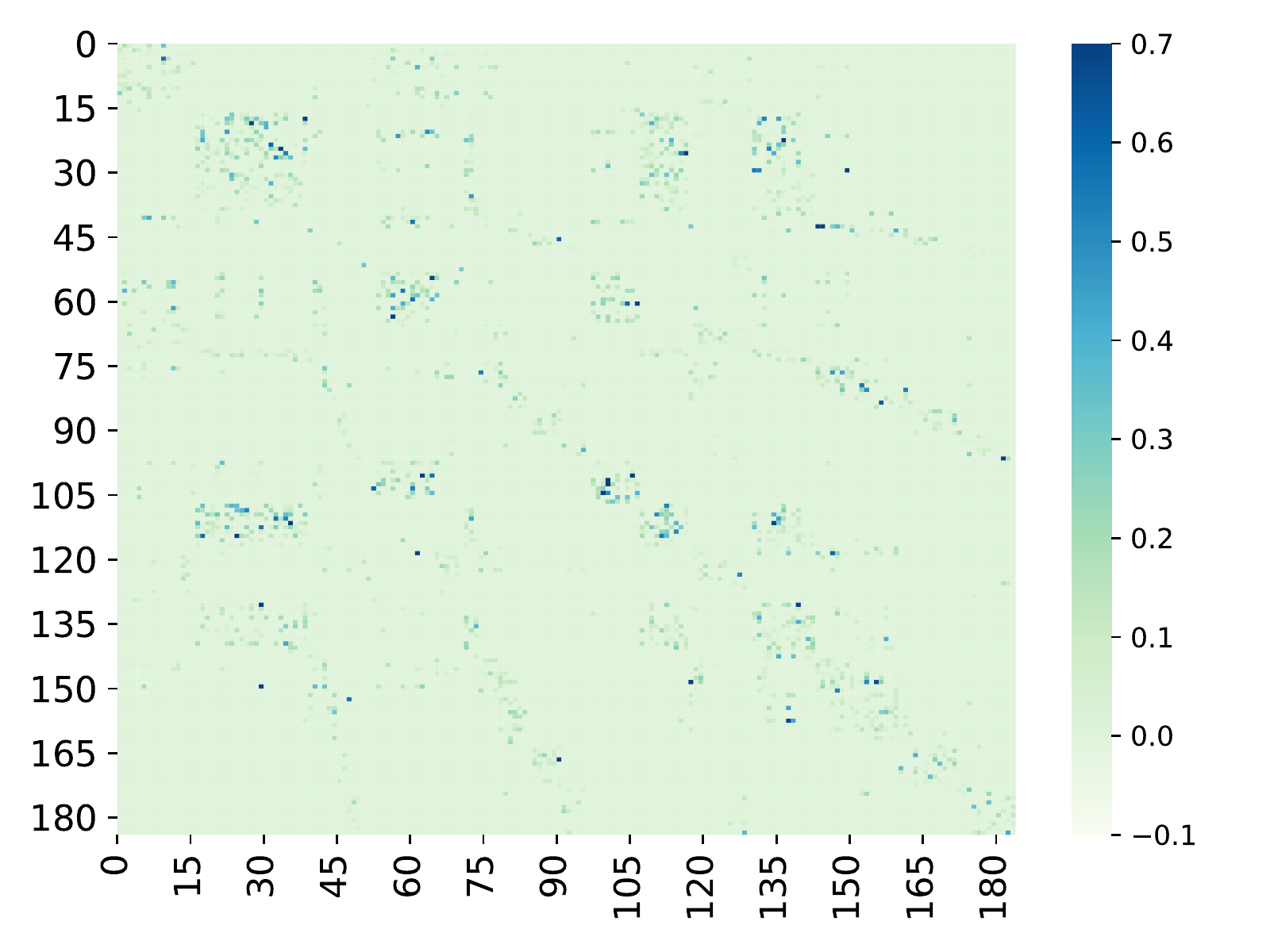}
    \label{pre6}
    \end{minipage}}
    \subfigure[The transpose of the upper triangle matrix of the adaptive edge attributes matrix minus the value of the lower triangle.]{
      \begin{minipage}{0.32\textwidth}
     \includegraphics[width=1\textwidth]{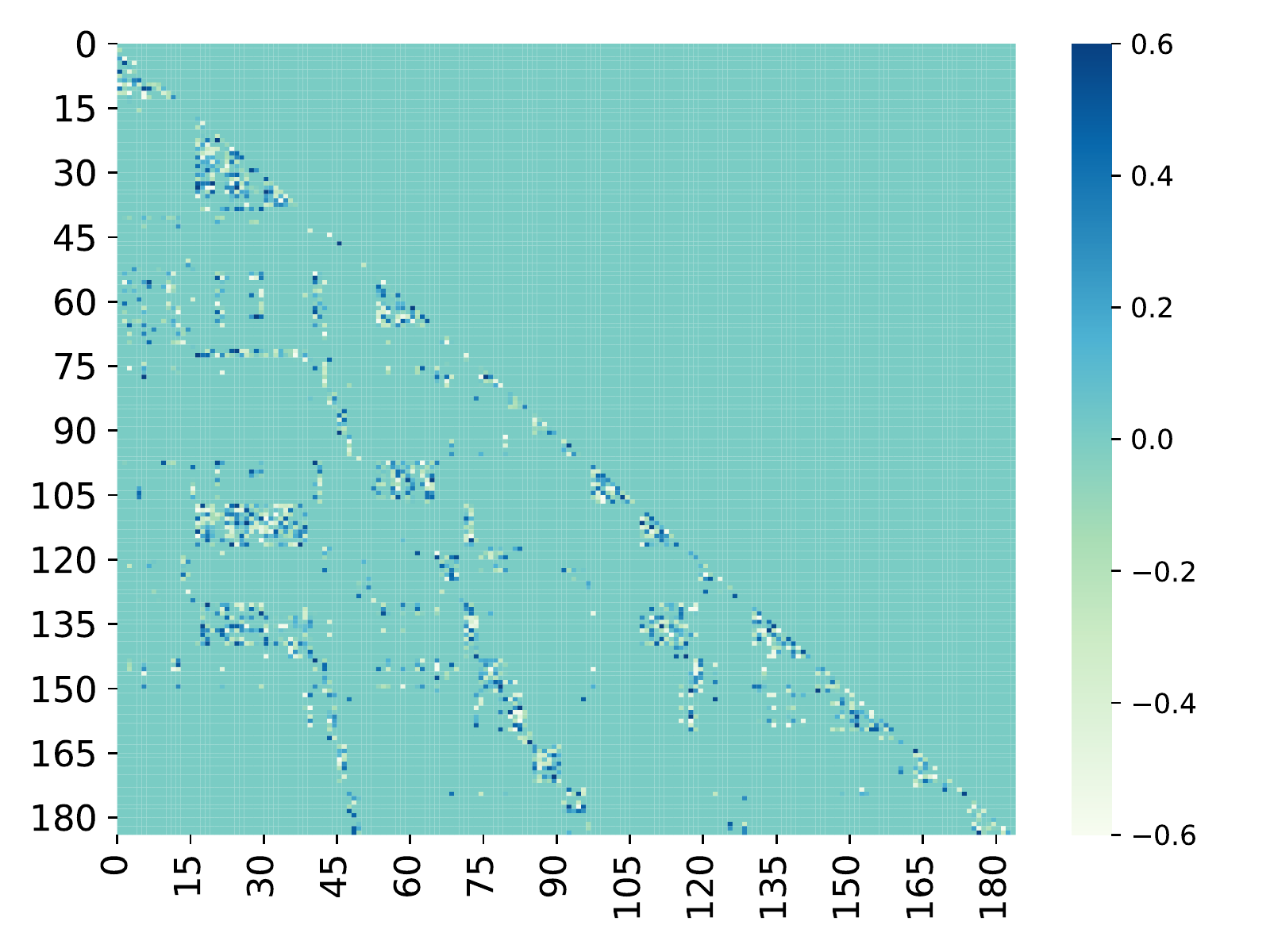}
     \label{pre12}
     \end{minipage}}
\caption{Difference between the two matrices use in our model.}
    \label{edges}
\end{figure*}

\subsection{The function of future weather}

Our model uses future weather data as known for future air quality prediction. We compare DGN-AEA and DGN-AEA (w/o weather) models respectively. As shown in Figure~\ref{barplots}, by using future meteorological data, the prediction accuracy can be improved by 8.87\%, 11.76\%, 13.65\%, 15.15\% on RMSE and 9.13\%, 12.30\%, 15.54\%, 17.76\% on MAE. This shows that it is very useful to use future weather data.

We compared our results with another another adaptive graph model without future weather, Graph WaveNet. Graph WaveNet requires a large number of parameters and has a high computational resource overhead, so the training is slow. Whats more, because the TCN model used by the time series module of Graph WaveNet can only perform convolution operations on one-dimensional variables, it cannot input future weather data. Compared with Graph WaveNet, our proposed model DGN-AEA(w/o weather) improves the RMSE of dataset 3 by 42.35\%, 38.25\%, 35.80\%, 33.98\%, and 31.86\%, 27.10\%, 26.13\%, 23.39\% on MAE. We speculate that it may be due to the construction of the adjacency matrix that changes from time to time brings great difficulty to training on $\mathrm{PM}_{2.5}$ datasets, making it difficult for the model to grasp the exact topology of the stations.

\subsection{Comparison of two attributes}

Following the above steps, we can get two kinds of graph edge attributes: one can be calculated by wind field data, and another can be learned after training. To get through the difference and explain why the adaptive edge attribute is useful, we visualize both of them at the same time step shown in Figure~\ref{edges}.

\subsection{Complex network analysis on the learned adaptive edges}
With DGN-AEA, an adaptive correlation network structure can be obtained after the training phase. The properties of the obtained correlation network can be discussed with the help of some analytical methods and indicators in the field of complex networks. We listed in the supplementary information.

We separately count the sum of the weights on the incoming and outgoing edges of different nodes and calculate the total weight of the connected edges minus the total weight of the incoming and outgoing edges. The positive or negative value of this difference indicates that the node belongs to the type that is more affected by the surroundings or has a greater influence on the surrounding (Figure~\ref{cities}). At the same time, we can also compare the relationship between edge weights and degree (Equation~\ref{E1}), and the relationship between edge weights and the degree centrality (Equation~\ref{E2}) of the complex network. Shown in Figure~\ref{centrality} and Figure~\ref{degree}. We see a clear positive correlation between node weight and degree value. However, there is no obvious correlation between node degree centrality and degree value.

\section{Conclusions}
\label{Conclusions}
In this paper, we propose a flexible Dynamic Graph Neural Networks with Adaptive Edge Attributes (DGN-AEA) based on the spatial domain. This method retains edges by the wind to follow the basic prior physical knowledge of air pollution transmission. At the same time, we calculate the transmission volume on the outgoing edge and incoming edge respectively when doing message transmission and aggregation to the nodes, which simulates the law of conservation of matter in the transport and diffusion of pollutants to some extent. Besides, we fuse adaptive edge attributes by means of the multi-graph structure. Experiment results show that our model achieves a better level of prediction effect on the real world $\mathrm{PM}_{2.5}$ dataset. 

In this way, we can adaptively learn the correlation between real sites and obtain better time series prediction results. However, how much of the network relationship in the real world can be restored in real data sets by this way of adaptively constructing learnable parameters is also worth exploring. There are still some ideas of network reconstruction methods that may be worth learning from.

\section*{Acknowledgment}
This work was supported by the National Natural Science Foundation of China (NO.42101027). We thank the support from the Save 2050 Program jointly sponsored by Swarma Club and X-Order.

\appendix
\section{Related Work about Air Quality Prediction}
Air quality prediction issues have been studied for years. These issues were first studied by some conventional statistical methods, e.g., autoregressive integrated moving average (ARIMA) \cite{yi2018deep,rekhi2020forecasting}. However, there are too much uncertainty and non-linearity in air quality prediction, which is not suitable for these statistical models to achieve high prediction accuracy for long-term prediction. 

Machine learning methods make use of historical observations to perform accurate predictions. Liu et al. \cite{liu2017urban} proposed a multi-dimensional collaborative support vector regression (SVR) model for air quality index (AQI) forecasting in the Jingjinji region while considering the weather conditions. Dun et al. \cite{dun2020short} adopted the linear regression (LR) and SVR method for short-term air quality prediction. Liu et al. \cite{liu2021analysis} fused the principal component regression (PCR), SVR, and autoregressive moving average (ARMA) models to predict the air quality with six different kinds of pollutants. However, these machine learning methods did not capture the spatio-temporal correlations and thus limiting the prediction performance.

In recent years, deep learning methods are widely employed in air quality prediction issues due to their high prediction accuracy. Ma et al. \cite{ma2020air} propose a transfer learning-based stacked bidirectional long short-term memory (LSTM) model which combined deep learning and transfer learning strategies to predict the air quality of some stations based on the data observed by other stations. Wen et al. \cite{wen2019novel} proposed a spatiotemporal convolutional long short-term memory neural network to capture the temporal and spatial dependencies with LSTM and convolutional neural networks (CNNs), respectively. Zhang et al. \cite{zhang2020pm2} proposed a hybrid model (MTD-CNN-GRU) for $\mathrm{PM}_{2.5}$ concentration prediction. In the MTD-CNN-GRU model, the CNNs were employed to extract the spatial relationships and the gated recurrent units (GRUs) were applied to capture temporal features. In this way, they could capture the spatio-temporal correlations to achieve higher prediction accuracy.

\section{Related Work about Graph-based Prediction Methods}
Conventional deep learning methods are not suitable for data processing in non-Euclidean space, which can not model the spatial correlations very well. To solve the problem, graph-based deep learning methods are proposed and have been widely applied to air quality forecasting these years. Wang et al. \cite{wang2021modeling} proposed an Attentive Temporal Graph Convolutional Network (ATGCN) for air quality prediction. The ATGCN encoded three types of relationships among air quality stations including spatial adjacency, functional similarity, and temporal pattern similarity into graphs and aggregated features using gated recurrent units (GRUs). Finally, a decoder was designed to conduct multi-step predictions. Qi et al. \cite{qi2019hybrid} then proposed a GC-LSTM model which combined the graph convolutional networks (GCNs) and LSTM to capture spatial terms and temporal attributes and predict the future $\mathrm{PM}_{2.5}$ concentrations. Wang et al. \cite{wang2020pm2} proposed a $\mathrm{PM}_{2.5}$-GNN model, which incorporated the domain knowledge into graph-structure data to model long-term spatio-temporal dependencies, for $\mathrm{PM}_{2.5}$ concentrations prediction. Since multiple features were considered, this model could achieve excellent prediction performance, especially for long-term predictions.

\section{Related Work about Dynamic Graph Models}
Recently, to better model contextual information, dynamic graph models have been employed by some researchers. Zhou et al. \cite{zhou2021forecasting} modeled a dynamic directed graph based on the wind field among the air quality stations. They then used the GCNs to capture the dynamic relationships among the stations and applied a temporal convolutional network (TCN) to predict the $\mathrm{PM}_{2.5}$ concentrations. Diao et al. \cite{diao2019dynamic} employed a dynamic Laplacian matrix estimator to model the dynamic graph, which can better model the spatial dependencies. Based on the dynamic estimator, they proposed a dynamic spatio-temporal graph convolutional neural network for traffic forecasting and outperformed the baselines. Peng et al. \cite{peng2021dynamic} employed the reinforcement learning to generate dynamic graphs and combined the graphs with the LSTM model for long-term traffic flow prediction. They further proved that dynamic graphs reduced the effects of data defects with extensive experiments.

\section{Related Work about Adaptive Graph Learning Models}

To overcome the limitations of prior information, Wu et al. \cite{wu2019graph} developed an adaptive dependency matrix through node embedding to capture the hidden spatial dependency in the data. And the model Multivariate Time Series Forecasting with Graph Neural Networks (MTGNN) \cite{wu2020connecting} also used this method to extract the uni-directed relations among variables. However, this method of changing the adjacency matrix with the time of the event will bring a lot of interference information to the training of the model, thereby affecting the accuracy of the prediction. Therefore, Graph WaveNet does not perform well on real air quality prediction datasets.

\section{Figures and Equations}
\begin{figure}[ht!]
    \centering
    \includegraphics[width=9cm]{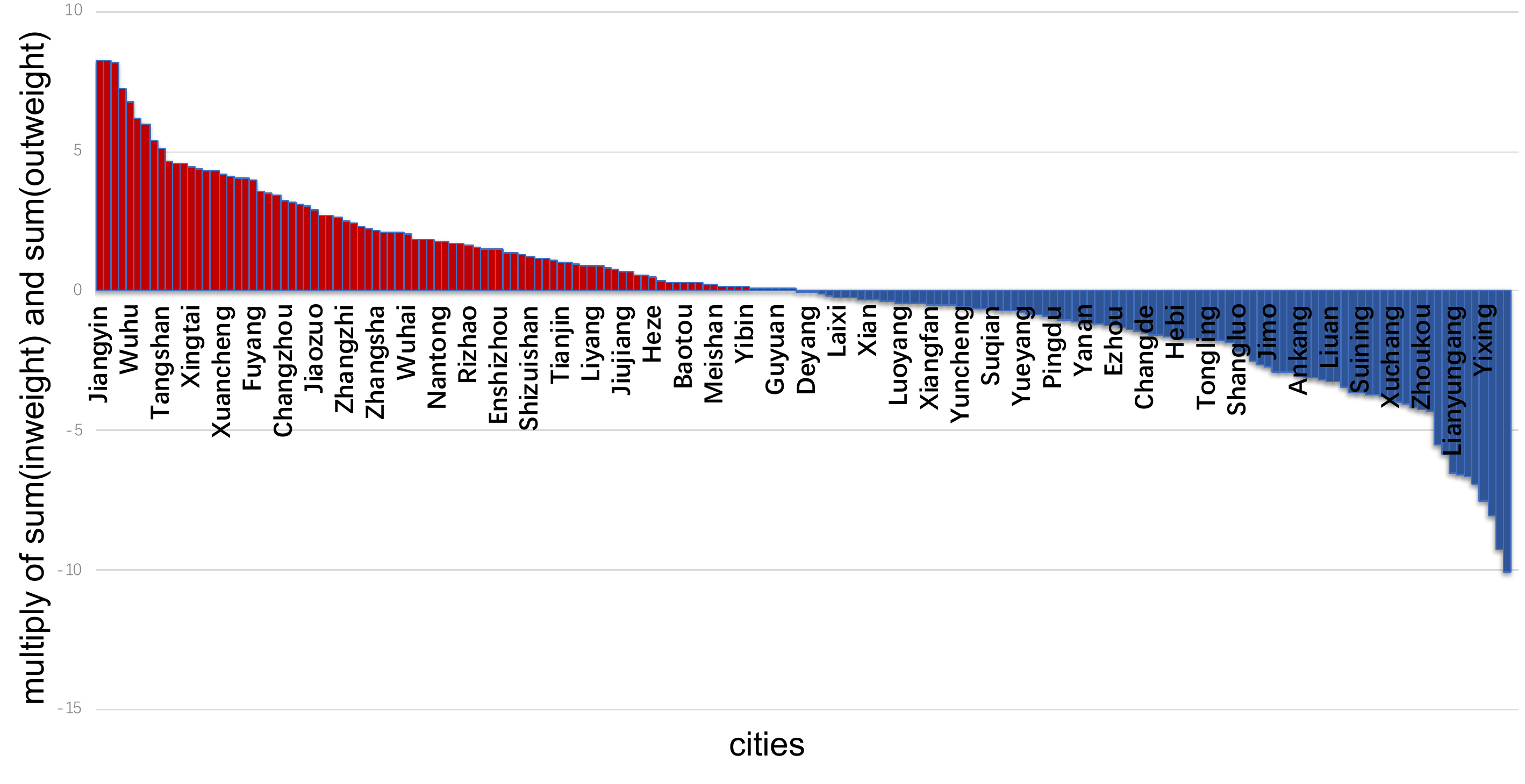}
    \label{cities}
    \caption{Cities sorted by the multiply of sum(in-weight) and sum(out-weight). The left red colored cities means they are attend to inflect their neighbors, and the right blue ones are just the opposite.}
\end{figure}

\begin{figure}[ht!]
    \centering
    \includegraphics[width=7cm]{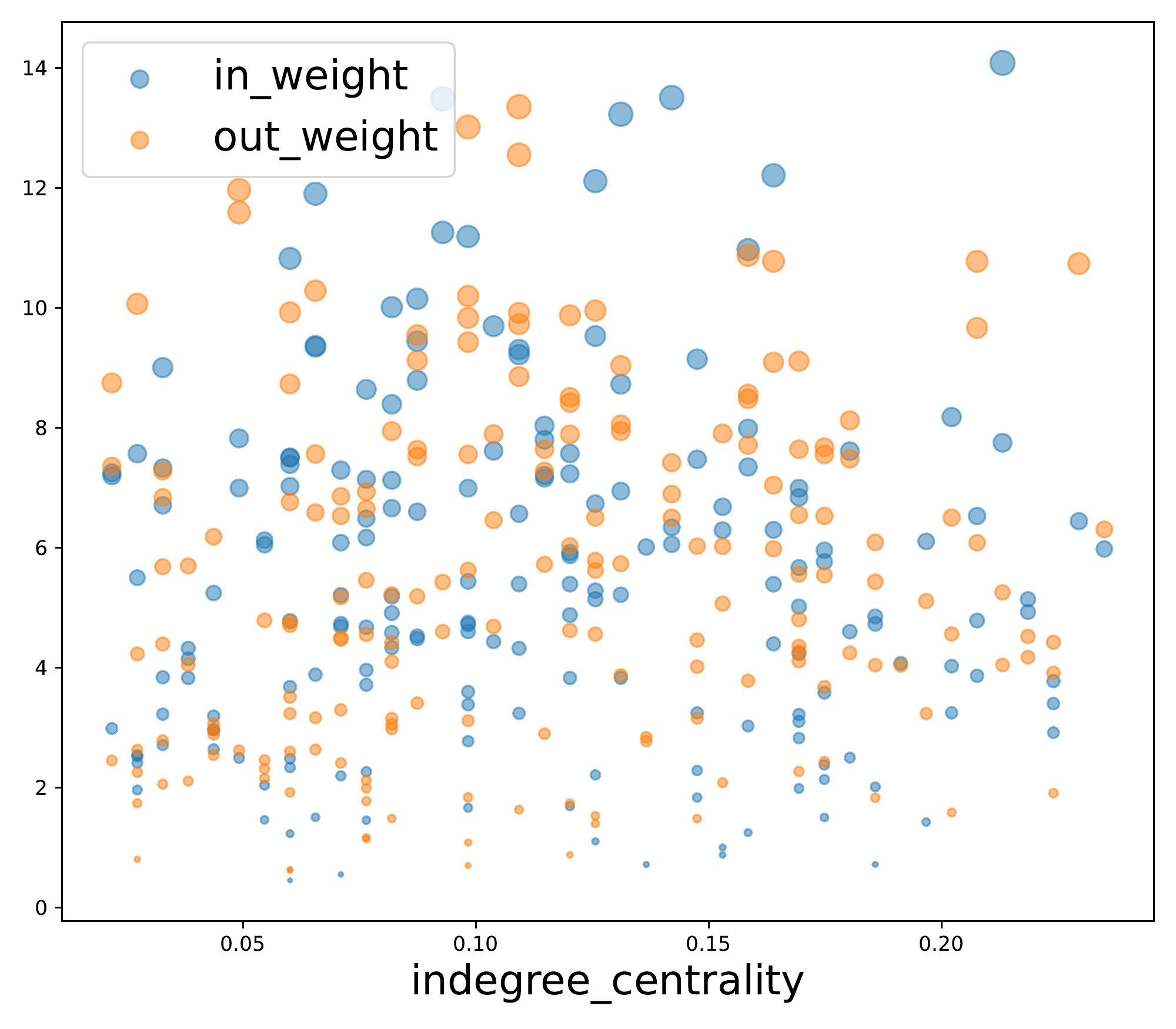}
    \caption{Relationships between degree centrality and connection weights.}
\label{centrality}
\end{figure}

\begin{figure}[ht!]
    \centering
    \includegraphics[width=7cm]{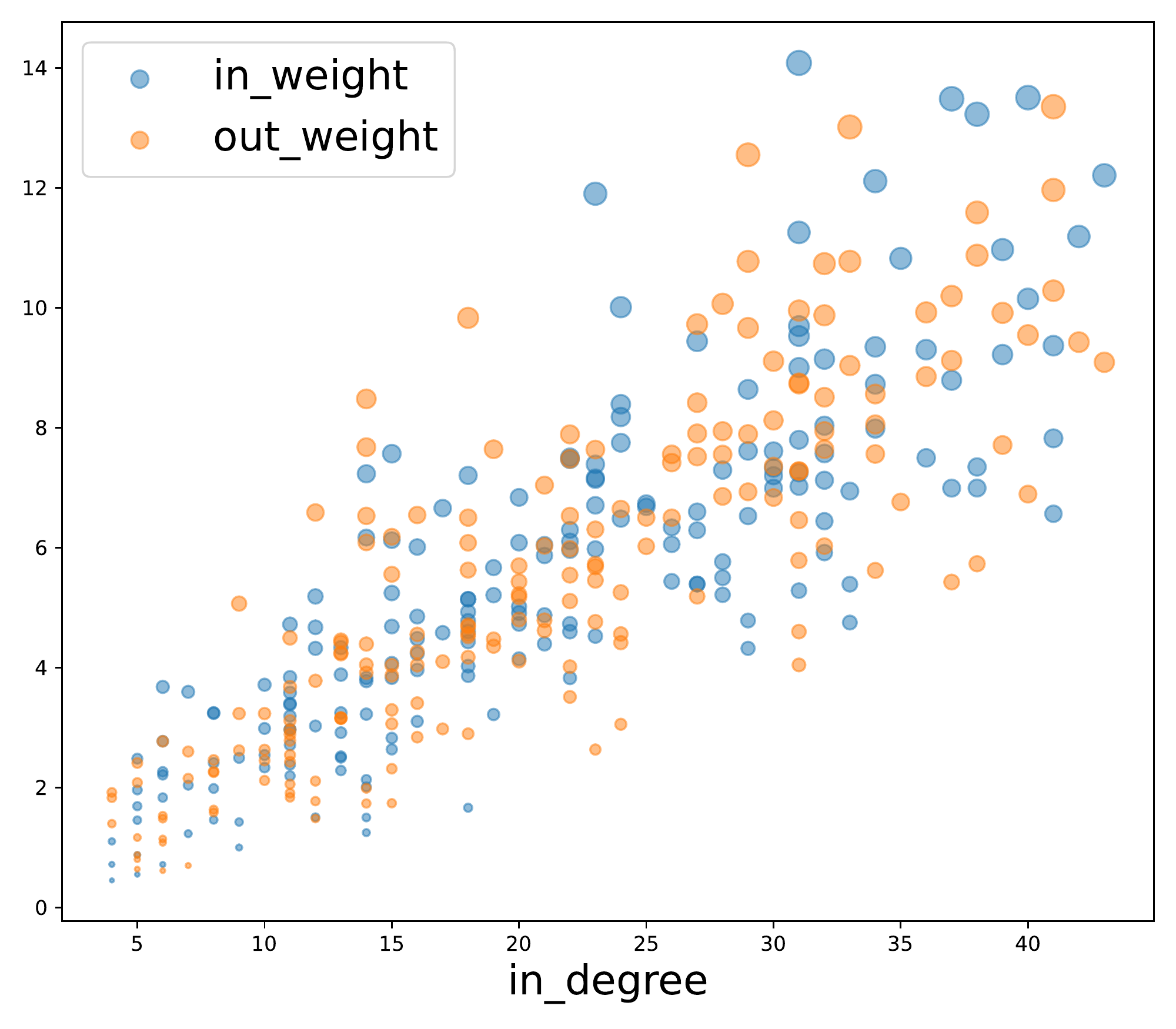}
    \caption{Relationships between degree and connection weights.}
\label{degree}
\end{figure}

\begin{equation}
    C_D\left(N_i\right)=\sum_{J=1}^{g}{x_{ij}\left(i\neq j\right)}
\label{E1}
\end{equation}

\begin{equation}
    {C\prime}_D\left(N_i\right)=\frac{C_D\left(N_i\right)}{g-1}
\label{E2}
\end{equation}

\bibliographystyle{elsarticle-num} 
\bibliography{air}


\end{document}